\documentclass{article}

    \usepackage[preprint]{neurips_2021}

\usepackage[utf8]{inputenc} 
\usepackage[T1]{fontenc}    
\usepackage{hyperref}       
\usepackage{url}            
\usepackage{booktabs}       
\usepackage{amsfonts}       
\usepackage{nicefrac}       
\usepackage{microtype}      
\usepackage{xcolor}         
\usepackage{hyperref}
\hypersetup{
    colorlinks=true,
    }
\usepackage{wrapfig}
\usepackage{times}
\usepackage{xspace}
\usepackage{epsfig}
\usepackage{amsmath}
\usepackage{amssymb}
\usepackage{graphicx}
\usepackage{mathtools}

\usepackage{booktabs} 
\usepackage{longtable}
\usepackage{enumitem}

\usepackage{url}
\usepackage{multicol}

\usepackage{caption}
\usepackage[skip=0.5ex]{subcaption}

\makeatletter
\DeclareRobustCommand\onedot{\futurelet\@let@token\@onedot}
\def\@onedot{\ifx\@let@token.\else.\null\fi\xspace}

\def\eg{\emph{e.g}\onedot} 
\def\ie{\emph{i.e}\onedot}

\def\etal{\emph{et al}\onedot}
\makeatother

\newcommand{\head}[1]{\noindent\textbf{#1}}

\newcommand\inv[1]{#1\raisebox{1ex}{$\scriptscriptstyle-\!1$}}

\DeclareMathOperator*{\argmin}{arg\,min}

\title{Residual Networks as Flows of Velocity Fields for Diffeomorphic Time Series Alignment}

\author{%
  Hao Huang \\
  New York University Abu Dhabi \\
  \texttt{hh1811@nyu.edu} \\
   \And
   Boulbaba Ben Amor \\
   Inception Institute of Artificial Intelligence \\
   \texttt{boulbaba.amor@inceptioniai.org} \\
   \AND
   Xichan Lin \\
   New York University \\
   \texttt{xl3417@nyu.edu} \\
   \And
   Fan Zhu \\
   Inception Institute of Artificial Intelligence \\
   \texttt{fan.zhu@inceptioniai.org} \\
   \And
   Yi Fang\thanks{Corresponding author} \\
   New York University Abu Dhabi \\
   \texttt{yfang@nyu.edu} \\
}

\begin{document}

\maketitle

\begin{abstract}
Non-linear (large) time warping is a challenging source of nuisance in time-series analysis. In this paper, we propose a novel diffeomorphic temporal transformer network for both pairwise and joint time-series alignment. Our ResNet-TW (Deep Residual Network for Time Warping) tackles the alignment problem by compositing a flow of incremental diffeomorphic mappings. Governed by the flow equation, our Residual Network (ResNet) builds smooth, fluid and regular flows of velocity fields and consequently generates smooth and invertible transformations (i.e. diffeomorphic warping functions). Inspired by the elegant Large Deformation Diffeomorphic Metric Mapping (LDDMM) framework, the final transformation is built by the flow of time-dependent vector fields which are none other than the building blocks of our Residual Network. The latter is naturally viewed as an Eulerian discretization schema of the flow equation (an ODE). Once trained, our ResNet-TW aligns unseen data by a single inexpensive forward pass. As we show in experiments on both univariate (84 datasets from UCR archive) and multivariate time-series (MSR Action-3D, Florence-3D and MSR Daily Activity), ResNet-TW achieves competitive performance in joint alignment and classification.
\end{abstract}

\section{Introduction}
\label{sec:intro}
Aligning temporal observations remains a challenging problem in time-series analysis \cite{srivastava2016functional}, space-time scene understanding in computer vision \cite{amor2015action}, behavioral analysis in medical imaging, and so on and so forth. Indeed, time-series data often presents a significant amount of misalignment, also known as non-linear time warping, which is usually caused by differences in execution or sampling rates. For instance, actions such as walking performed by different actors have different execution rates and different starting points in a periodic process owing to physiological and bio-mechanical factors. Veeraraghavan \etal~\cite{veeraraghavan2009rate} showed that ignoring such temporal variability can greatly decrease recognition performance. Temporal alignment seeks to find a plausible temporal transformation between a query sequence and a target sequence, such that their temporal variability is minimized (synchronizing observations in the previous walking example). In this work, we propose ResNet-TW, a novel diffeomorphic temporal transformer network for both pairwise and joint time-series alignment. Inspired by the geometric LDDMM (Large Deformation Diffeomorphic Metric Mapping) framework \cite{beg2005computing}, we introduce the flow equation as an additional constraint to govern the construction of the transformation on the basis of fluid flow of vector fields. To this end, we accommodate Deep Residual Networks to predict and integrate such flows of non-stationary vector fields. Then, similarly to \cite{weber2019diffeomorphic}, we formulate the joint alignment problem as to simultaneously compute the centroid and align all sequential data within a class, under a semi-supervised schema. 

\head{Contributions and paper's organization.} \textbf{1)} We restate training Residual Networks as integrating non-stationary velocity fields in temporal LDDMM to compute/learn warping functions for time-series alignment. A similar interpretation of ResNets as incremental flows of diffeomorphisms was recently presented in \cite{rousseau2019residual} in the context of supervised learning with application to image classification. \textbf{2)} We propose ResNet-TW, a diffeomorphic temporal transformer network for both pairwise and joint alignment of time-series. Compared to existing transformers (\eg DTAN \cite{weber2019diffeomorphic} and TTN \cite{lohit2019temporal}), ResNet-TW guarantees diffeomorphic warping under large misalignment, \textbf{3)} We conduct extensive experiments on several publicly available datasets to validate the generalization ability of our models to unseen data for time-series joint alignment and classification. The rest of our paper is organized as follows: In Section \ref{sec:related}, we provide a mathematical formulation of both pairwise and joint alignment problems in time-series analysis and review existing works. Section \ref{sec:propre} describes the proposed approach. Implementation details and experimental evaluations are reported in Section \ref{sec:exp}. Section \ref{sec:Conclusion} provides some concluding remarks and opens some perspectives.

\section{Problem formulation and related work}
\label{sec:related}

\head{Pairwise alignment.} Given two time-series $f$ and $g: \Omega \to \mathbb{R}^d$, observations of the same temporal process where $\Omega \subseteq \mathbb{R}$ is the time domain, \ie, a time interval $[\tau_1, \tau_2]$, on which the sequential data are defined, the key problem is to find an optimal (plausible) warping function $\gamma^\ast:\Omega \to \Omega$, such that the data fit between $f$ and $g \circ \gamma^\ast$ is high (\ie, their discrepancy according to a distance measure ${\mathcal D}$ is minimum). This problem is typically solved as an optimization problem~\cite{srivastava2016functional}, as in Eq. (\ref{eq:pairwise_problem}),   
\begin{equation}
    \gamma^\ast = \argmin_{\gamma \in \Gamma}{\mathcal{D}}(f, g \circ \gamma) + {\mathcal{R}}(\gamma) \enspace.
    \label{eq:pairwise_problem}
\end{equation}
While the first data term evaluates the similarity between $f$ and $g \circ \gamma$, $\gamma \in \Gamma$, and the second regularization imposes constraints on the warping function $\gamma$, as smoothness, monotonicity preserving and boundary conditions. The set $\Gamma$ is then the set of all monotonically-increasing functions from $\Omega$ to itself with $\gamma(0)=0$ and $\gamma(1)=1$. In time-series analysis, it is desirable to have a warping function $\gamma$ that is invertible, and both $\gamma$ and $\gamma^{-1}$ (its inverse) sufficiently smooth, \ie $\gamma$ is a diffeomorphism.

\head{Joint alignment.} The goal of joint time-series alignment is to find an \textit{average temporal sequence} $\bar{g}$ for a given set of $N$ time-series observations $G=\{g_1, g_2, \cdots, g_N\}$ in a space $E$ induced by a distance measure $\mathcal{D}$, such that $\bar{g}$ minimizes the sum of distances to all elements in the set $G$ (Eq. \ref{eq:avg_obj}), 
\begin{equation}
    \bar{g} = \argmin_{g \in  E} \sum_{i=1}^N \mathcal{D} (g, g_i)\enspace.
    \label{eq:avg_obj}
\end{equation}
In analogy to Eq. (\ref{eq:pairwise_problem}), the joint alignment can be solved by finding a set of optimal warping functions $\{\gamma_i^\ast\}_{i=1}^N$ verifying Eq. (\ref{eq:joint_problem}),
\begin{equation}
    \{\gamma_i^\ast\}_{i=1}^N = \argmin_{\{\gamma_i \in \Gamma\}_{i=1}^N} \sum_{i=1}^N \big[{\mathcal{D}}(\bar{g}, g_i \circ \gamma_i) + {\mathcal{R}}(\gamma_i)\big] \enspace,
    \label{eq:joint_problem}
\end{equation}
where each optimal warping function $\gamma_i^\ast:\Omega \to \Omega$ minimizes the discrepancy between $g_i$ and $\bar{g}$. Solve Eq. (\ref{eq:joint_problem}) jointly computes a consensual time-series and align all times-series to it.  


\head{Dynamic Time Warping (DTW) and variants.} A popular approach for pairwise alignment of time-series is DTW~\cite{sakoe1971dynamic,sakoe1978dynamic} which, by solving the Bellman’s recursion via dynamic programming~\cite{bellman1959adaptive}, finds an optimal monotonic alignment between two time-series. The time cost for DTW to align two time-series of respective length $n$ and $m$ is $\mathcal{O}(mn)$. In~\cite{cuturi2017soft} Cuturi \etal have defined Soft-DTW, a differentiable loss function, that can be integrated into neural networks. It is also used for averaging time series using DTW discrepancy~\cite{cuturi2017soft}. However, it is only capable of aligning currently-available data, and compute from scratch for new data, which is especially computation-consuming when the new data is much larger than the original one. Srivastava \etal have derived in \cite{srivastava2010shape} an interesting representation, termed SRVF (for Square Root Velocity Function) and an elastic metric that is invariant to the reparameterization group. Optimal warping functions are computing using Dynamic Programming. In \cite{amor2015action}, Ben Amor \etal accommodated this representation to skeletal trajectories on the Kendall's shape space for action recognition. Advanced variants of DTW such as Canonical Time Warping (CTW) \cite{zhou2009canonical} finds the optimal reduced dimensionality subspace such that the sequences are maximally linearly correlated. Generalized Time Warping (GTW) \cite{zhou2015generalized} uses a combination of CTW and a Gauss-Newton temporal warping method that parameterizes the warping path as a combination of monotonic functions. Deep learning versions of CTW have also been recently proposed~\cite{trigeorgis2016deep}.


\head{Temporal Transformers Networks.} Following the spatial transformer methodology proposed first in \cite{jaderberg2015spatial}, several temporal transformers nets have been proposed recently. Among them, the Diffeomorphic Temporal Alignment Nets (or DTAN) and its recurrent variant (R-DTAN) are recently proposed in \cite{weber2019diffeomorphic} for pairwise and joint time-series alignment. DTAN learns and applies an input-dependant transformation to the input signal to minimize a loss function, including a regularizer. That is, in single-class problems, DTAN turns to an unsupervised method for registration (\ie solve Eq. (\ref{eq:pairwise_problem})), while in multi-class case, it yields into a semi-supervised approach where class labels are considered (\ie solve Eq. (\ref{eq:joint_problem})). To deal with large misalignment, a recurrent variant of DTAN is also proposed. R-DTAN guarantees diffeomorphic transformations by incrementally composing smaller diffeomorphic transformations, however, this aspect was omitted in \cite{weber2019diffeomorphic}. To reduce the number of its parameters, R-DTAN integrates stationary velocity fields by sharing the learned parameters by all the temporal transformer layers. Taking a different direction, in \cite{lohit2019temporal} a separate temporal transformer was integrated at the front end of a classification network. This temporal transformer network (TTN) lead to rate-robust representations that reduce intra-class variability. In \cite{nunez2020deep} a deep architecture for learning warping functions is proposed for elastic shape analysis (using SRVF representation as in \cite{srivastava2010shape} and \cite{amor2015action}). A temporal transformer network (TTN) is trained on shape representations derived from the original shapes then used to predict optimal warping functions separating unseen shapes. Except for R-DTAN, approaches cited above belong to the family of \textit{Elastic Models} in which warping functions $\gamma: \Omega \to  \Omega$ are generated by perturbations from the identity, i.e. $\gamma(t)=Id(t)+v^{\theta}(t)$, $t\in \Omega$ and $\theta$ are parameters of the network. Consequently, they can not guarantee diffeomorphisms for large misalignment (see \cite{beg2005computing} for further argumentation). 

\head{Consensus sequence problem (Joint alignment).} In the context of statistical inference and time-series classification, many algorithms require a method to represent information from a set of objects in a \textit{sample average}. DTW is intractable for large-scale joint alignment problems due to a quadratic time cost for pairwise alignment. Averaging under the DTW distance is a nontrivial task, as it involves solving the joint-alignment problem. A congealing algorithm solves iteratively for the joint alignment by gradually aligning one signal towards the rest~\cite{learned2005data}. Typical alignment criteria used in congealing are entropy minimization~\cite{miller2000learning,huang2007unsupervised} or least squares~\cite{cox2008least,cox2009least}. Dalca \etal \cite{dalca2019learning} proposed a learning-based method for building deformable conditional templates based on diffeomorphisms. While several authors proposed smart solutions for the averaging problem~\cite{sun1993time,petitjean2011global,cuturi2014fast,petitjean2014dynamic,weber2019diffeomorphic}, none of them except for~\cite{weber2019diffeomorphic} does not require solving a new optimization problem each time for aligning new sequential data. Recently, ~\cite{lohit2019temporal} proposed a temporal transformer network based on 1D diffeomorphisms for time series classification, 
but does not scale well with the signal’s length. ~\cite{detlefsen2018deep} showed it is possible to explicitly incorporate flexible and efficient diffeomorphisms~\cite{freifeld2015highly,freifeld2017transformations} within deep learning models via a Spatial Transformer Network (STN)~\cite{jaderberg2015spatial}; particularly, they focused on supervised learning for image recognition and classification. 

\section{Proposed approach}
\label{sec:propre}

Registration problems, including temporal alignment, have been widely stated as the estimation of diffeomorphic transformations between input and output data. By parameterizing the transformation via integration of sufficiently regular stationary or time-dependent velocity fields~\cite{beg2005computing,vercauteren2009diffeomorphic,hart2009optimal,chen2013large} and imposing sufficient regularization, diffeomorphic transformations can be assured. This was the central idea of \textit{Large Deformation Diffeomorphic Metric Mapping} (LDDMM)~\cite{beg2005computing} which tackles the registration issue through a composition of a series of incremental diffeomorphic mappings, each individual mapping being close to an identity mapping. More formally, we refer to $g$ as a query and $f$ as a target time-series. It is desirable to have the warping function $\gamma$ between $f$ and $g$ to be invertible, and both $\gamma$ and $\inv{\gamma}$ to be sufficiently smooth, \ie, $\gamma$ is a diffeomorphism. This set of diffeomorphisms forms a group with the identity mapping as the neutral element. As derived in~\cite{beg2005computing}, the transformation map is time-dependent and is generated as the result of the integration over time of smooth velocity field $v: [0,1] \times \Omega \to \Omega$, governed (or constrained) by the the \textit{Flow Equation} (Eq. (\ref{eq:flowEq})), 

\begin{equation}
    \dot{\gamma}(t, \tau) \coloneqq \frac{\partial \gamma(t, \tau)}{\partial t} = v(t, \gamma(t, \tau)), \gamma(0, \tau)=\tau\enspace,
    \label{eq:flowEq}
\end{equation}
for all $\tau \in \Omega$, $t \in [0, 1]$, and $v(t, \gamma(t, \tau))$ is a time-dependent velocity field. Accordingly, a time-dependent transformation $\gamma: [0, 1] \times \Omega \to \Omega$ is modeled where $\gamma(t, \tau) \in \Omega$ describes the position of a particle at time $t$ that was at time $0$ in $\tau \in \Omega$. The warping $\gamma(t, \tau)$ is obtained by integration of the flow equation as follows (Eq. (\ref{eq:lddmm_v2gamma})),

\begin{equation}
    \gamma(t, \tau) = \gamma(0, \tau) + \int_0^t v(s, \gamma(s, \tau)) ds \enspace.
    \label{eq:lddmm_v2gamma}
\end{equation}

This gives a path $\gamma_t=\gamma(t,.): \Omega \to \Omega, t \in [0, 1]$ in the space of warping functions starting with $\gamma_0=\gamma(0,.)$ as an identity mapping and terminating at the end-point $t=1$ of a particular warping function $\gamma(1,.)=\gamma_1 = \gamma_0 + \int_0^1 v_t(\gamma_t)dt$ transforming $g$ to $f$. Accordingly, Eq. (\ref{eq:pairwise_problem}) can be phrased as follows: find a flow of velocity fields $v^\ast: [0,1] \to \Omega$ as:
\begin{equation}
v^\ast = \argmin_{v:\dot{\gamma}_t = v_t(\gamma_t)} \Big( \mathcal{D}(f, g \circ \gamma_1) + \alpha \int_0^1  \lVert v_t \rVert _V^2 dt \Big)\enspace,
\label{eq:lddmm_goal}
\end{equation}
where $\alpha$ balances the weight between data and regularization terms. In LDDMM formulation, $V$ is an Reproducing Kernel Hilbert Space (or RKHS) defined by the Gaussian kernel to guarantee smoothness and regularity of computed velocity fields and $\lVert . \rVert _V$ is the induced RKHS norm. In addition, the quantity $\int_0^1  \lVert v_t \rVert _V^2 dt$ plays the role of a \textit{Kinetic energy} of the velocity field $v$ and allows to measure the length of the path separating $\gamma_0$ and $\gamma_1$. While the LDDMM framework have been widely used in shape and image registration, to our knowledge its extension to time-series (except for time series of images \eg \cite{hadj2016longitudinal}) is not explored yet. Furthermore, this work is among very few recent works (\eg \cite{rousseau2019residual}) to revise the LDDMM framework using Deep Residual Networks.    


\begin{wrapfigure}{r}{0.32\textwidth}
\includegraphics[width=\linewidth]{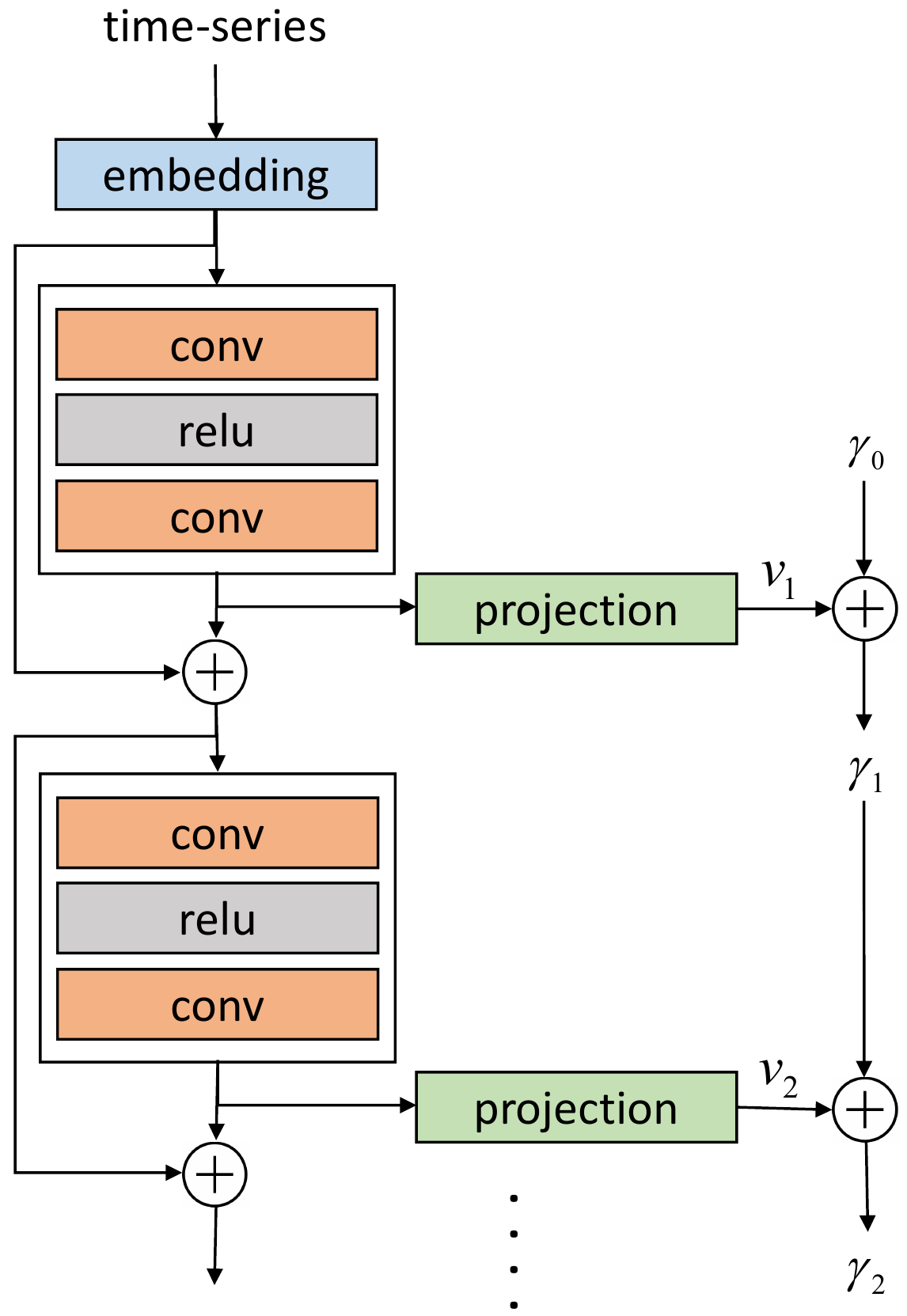} 
\caption{ResNet-TW for velocity fields prediction and integration to generate diffeomorphic warpings.}
\label{fig:model}
\end{wrapfigure}

\head{Eulerian discretization using deep residual networks.} Deep learning models have been applied for a wide range of problems. Among the proposed architectures, \textit{Residual Networks} (also called ResNets) have achieved state-of-the-art performance in supervised learning. In this paper, inspired by insights into ResNets from an aspect of ordinary/partial differential equation (ODE/PDE)~\cite{haber2017stable,ruthotto2019deep,weinan2017proposal}, we regard ResNets as numerical schemes of differential equations and relate the incremental mapping defined by ResNets to diffeomorphic registration models, especially to \textit{Large Deformation Diffeomorphic Metric Mapping} (LDDMM) \cite{rousseau2019residual}. Concretely, in our ResNet-TW, the $l$-th residual block computes an update in the form of,
\begin{equation}
    \gamma_{l} = \gamma_{l-1} + F(\gamma_{l-1}; W_l). 
    \label{eq:res_block}
\end{equation}
where $\gamma_{l-1}$ is the input to the $l$-th residual block $F(.; W_l)$ ($\gamma_0$ is the identity mapping), and $W_l$ is a set of weights and biases associated with the $l$-th residual block. Specifically, the $l$-th residual block predicts the velocity field $v_l=F(\gamma_{l-1}; W_l)$ that is added to the warping function $\gamma_{l-1}$. By imposing regularity on the norm of $v$, which lies to the space of neural functions (assimilated to a RKHS \cite{bietti2017invariance}), the entire ResNet implements the composition of a series of incremental diffeomorphic mappings, which is a discretized version of Eq. (\ref{eq:lddmm_v2gamma}), \ie, by replacing integral with summation. An instantiation of ResNet-TW is shown in Figure~\ref{fig:model} which builds on three main steps (only two building blocks are illustrated),
\vskip -0.2in
\begin{itemize}
\item[--] An \textit{embedding} step consisting of a single convolutional layer which embeds the input time series data from an initial low-dimensional space to a higher dimensional space driven by the number of filters used.
\item[--] A series of identical \textit{residual blocks} $F(.; W_l)$ which computes time-dependent (non-stationary) velocity fields ($ W_l$ are different). In the core of each block, a point-wise \textit{ReLu} activation function is applied to introduce non-linearity. 
\item[--] A series of \textit{projection} operations (\ie dimensionality reduction) ends each of residual blocks and allow to cast estimated a velocity fields such that $v_l: \Omega \to \Omega$. Consequently, the outputs $\gamma_l$ are also $\Omega \to \Omega$ by summation of $v_l$ over the residual blocks $l\in[1,l]$ and the initial warping function $\gamma_0$. 
\end{itemize}

By constraining our temporal transformer to control the amount of \textit{kinetic energy} introduced by elementary velocity $v_l$ (\ie the network activity), we guarantee (1) diffeomorphic intermediate $\gamma_l$ and final warping functions $\gamma_1$, and (2) optimal warping functions, in terms of length of the path $\gamma_t\in \Gamma$ connecting $\gamma_1$ and $\gamma_0$. To our knowledge, ResNet-TW is the first temporal (transformer) alignment method that propose this solution inspired by the LDDMM framework \cite{beg2005computing}. We notice that unlike ResNet-TW, previous approaches (\eg  DTAN \cite{weber2019diffeomorphic} and TTN \cite{lohit2019temporal}) compute perturbations from the identity and thus do not guarantee diffeomorphic warping functions. As far R-DTAN (the recurrent version of DTAN \cite{weber2019diffeomorphic}) is concerned, it predicts stationary velocity fields (\ie a  soothing approach) which make the approach more efficient, by computing an initial velocity $v_0$, but less flexible than our ResNet-TW (\ie a relaxation approach). We will further illustrate these advantages in Section \ref{subsec:msr}. In single-class cases, this yields an unsupervised method for joint-alignment learning. In multi-class cases, this forms a semi-supervised method in which only class labels are used during training to align data from multiple categories.

\head{Loss function for multiple alignment.} Following the formulation in \cite{weber2019diffeomorphic}, the data term for single-class joint alignment is defined as Eq.~(\ref{eq:single-classLoss}),
\begin{equation}
    \sum_{i=1}^N {\mathcal{D}}(\bar{g}, g_i \circ \gamma_i) = \frac{1}{N}\sum_{i=1}^N \lVert \hat{g}_i - \frac{1}{N}\sum_{j=1}^N\hat{g}_j \rVert _{\ell_2}^2\enspace,
    \label{eq:single-classLoss}
\end{equation}
where $\hat{g}_i = g_i \circ \gamma_i$. Note this setting is unsupervised and $\bar{\hat{g}} = \frac{1}{N}\sum_{j=1}^N\hat{g}_j$ is the average sequence of the warped data. For multi-class joint alignment, the data term is the sum of the within-class variances (Eq.~(\ref{eq:multi-classLoss})),
\begin{equation}
   \sum_{i=1}^N {\mathcal{D}}(\bar{g}, g_i \circ \gamma_i) = \sum_{k=1}^K \frac{1}{N_k}\sum_{i:z_i=k} \lVert \hat{g}_i - \frac{1}{N_k}\sum_{j:z_j=k}^N\hat{g}_j \rVert _{\ell_2}^2\enspace,
   \label{eq:multi-classLoss}
\end{equation}
where $K$ is the number of classes, $z_i$ takes values in $\{1, \dots, K\}$ and is the class label associated with $g_i$, \ie, $z_i = k$ if and only if $g_i$ belongs to class $k$, and $N_k = \lvert \{i:z_i = k\}\rvert$ is the number of observations in class $k$. This is a semi-supervised setting as proposed in~\cite{weber2019diffeomorphic}. That is, the labels $\{z_i\}^N_{i=1}$ are available during the training, but not during the testing. Importantly, note that the same single network is responsible for aligning each of the classes, \ie, $W$ does not vary with $k$. Following \cite{weber2019diffeomorphic} and \cite{freifeld2017transformations}, the regularization term in Eq. (\ref{eq:joint_problem}) in respectively joint alignment is defined as,
\begin{equation}
    \mathcal{R}(\gamma) = \sum_{l=1}^L \mathbf{a}_l^\top \inv{\Sigma} \mathbf{a}_l\enspace,
    \label{eq:regularization}
\end{equation}
where $\mathbf{a}_l = [a_l^1, \cdots, a_l^{N_{\mathcal{T}}}]$ and $\Sigma$ is a zero-mean Gaussian with a $N_{\mathcal{T}} \times N_{\mathcal{T}}$ covariance matrix whose correlations decay with inter-cell distances. Notice that due to $v_l^i = a_l^i\tau + b_l^i$, if we choose $\Sigma$ as an identity matrix, Eq. (\ref{eq:regularization}) is the discretized version of the regularization term in Eq. (\ref{eq:lddmm_goal}). Similar to~\cite{freifeld2017transformations}, $\Sigma$ has two parameters: $\lambda_{var}$, which controls the overall variance, and $\lambda_{smooth}$, which controls the smoothness of velocity fields. A small $\lambda_{var}$ favors small warps (\ie, close to the identity) and vice versa; similarly, a large $\lambda_{smooth}$ favors velocity fields that are almost purely affine and vice versa. 

\begin{figure}[!ht]
\includegraphics[width=.5\linewidth]{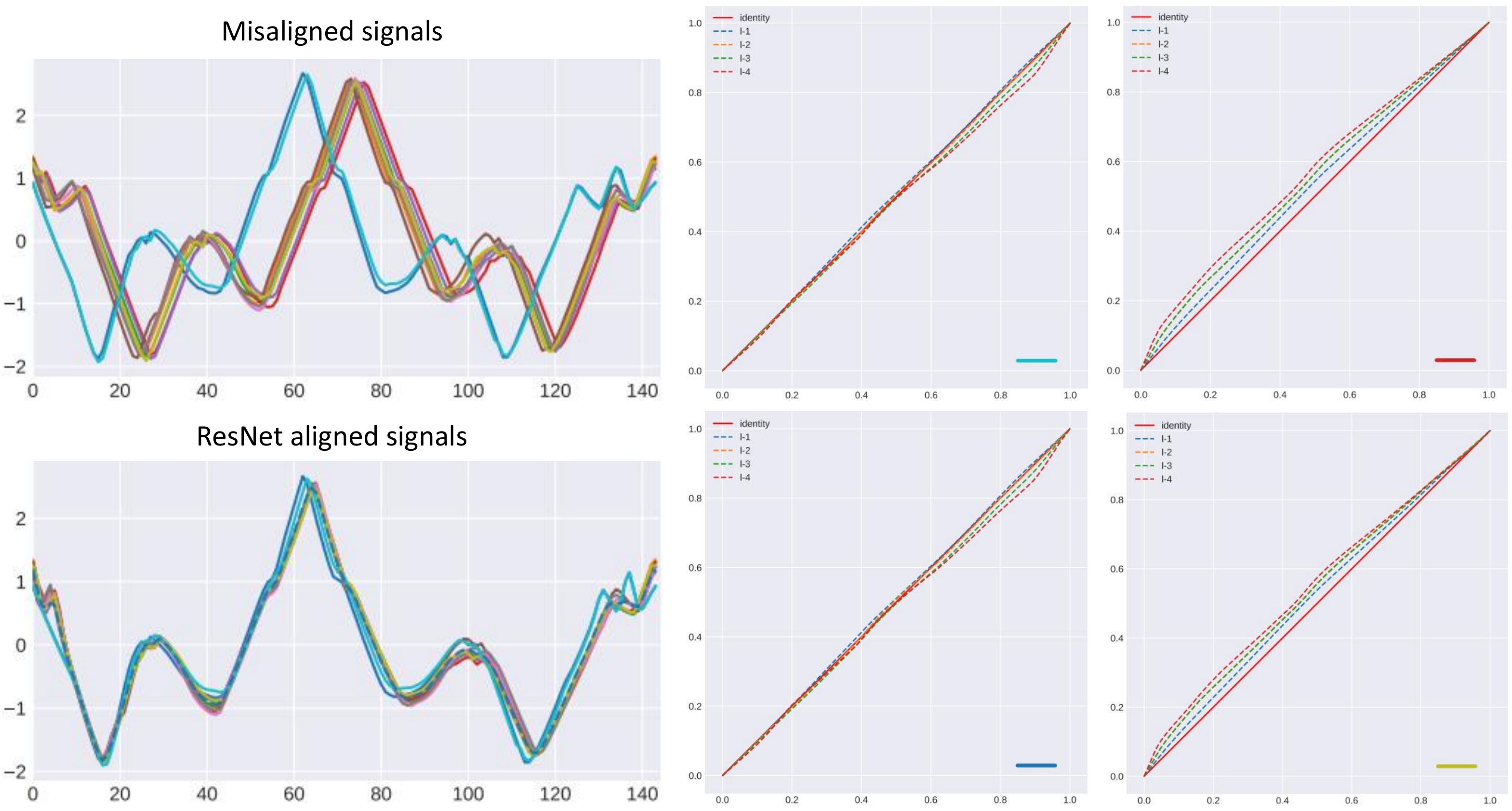}
\includegraphics[width=.5\linewidth]{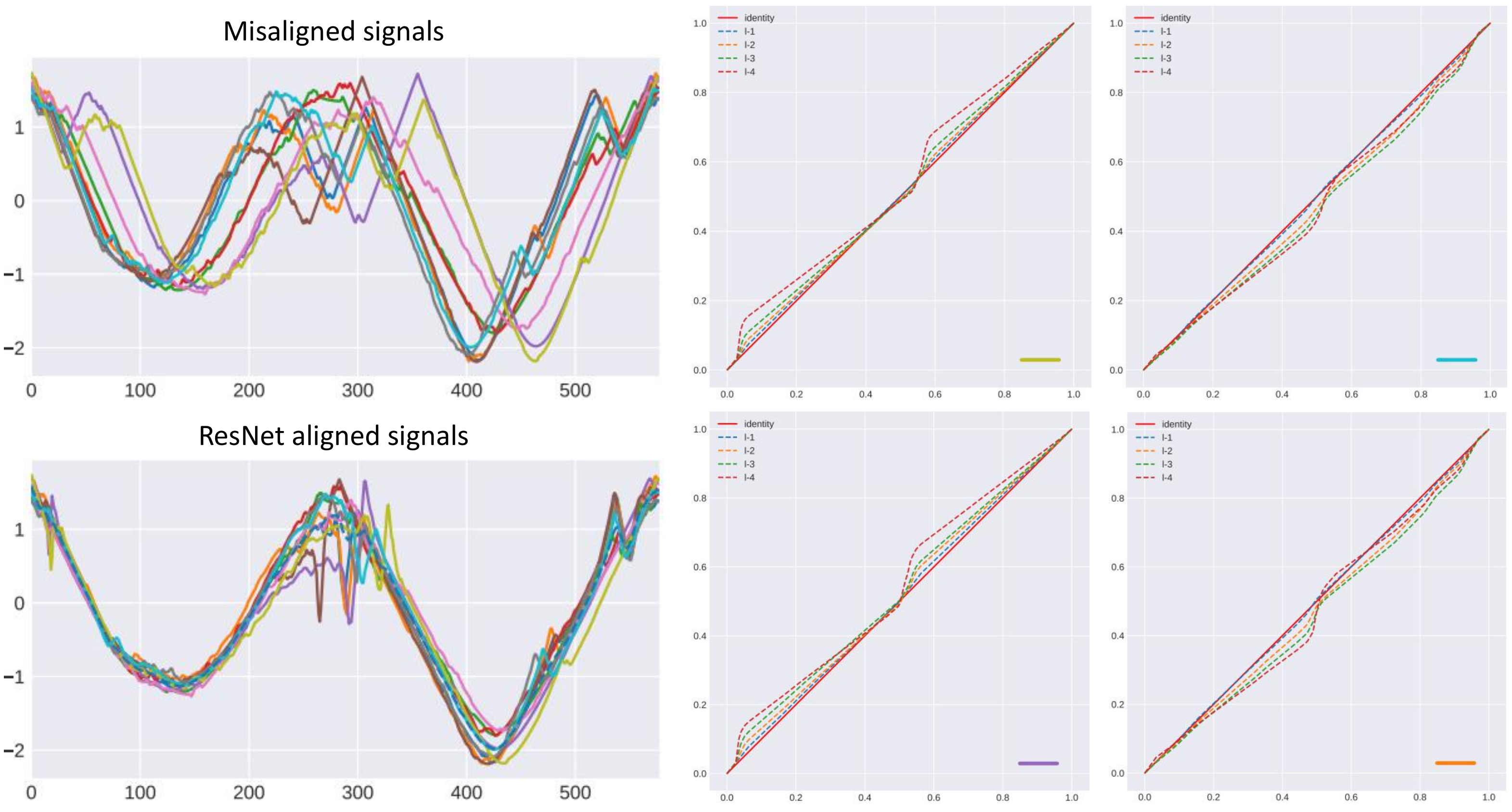}
\caption{Within-class joint alignment of two datasets (Plane and Car) from the UCR archive. The illustrated results are of previously unseen test samples. The colors of short bars at the right-bottom on the right part correspond to signals on the left part.}
\label{fig:ucr_align}
\end{figure}

In Figure \ref{fig:ucr_align}, we illustrate within-class joint alignment results of sequences taken from the UCR archive Plane and Car datasets. The illustrated results are of previously unseen test samples. Both misaligned and aligned data are shown. The colors of short bars at the right-bottom on the right part correspond to signals on the left part. (More examples are provided in our \textbf{supplementary materials}.)

\head{Monotonic constraint in Time Warping (TW).} We focus on a specific set $\Gamma$ of warping functions. Given a 1-differentiable function $\gamma_t$ obtained at any $t \in [0, 1]$ and defined on the domain $\Omega = [0, 1]$, for $\gamma_t$ to be an element of $\Gamma$, $\gamma_t$ needs to satisfy the following conditions\footnote{The second $[0, 1]$ has different meaning with the first $[0, 1]$.}:
\begin{equation}
    \begin{cases}
        \gamma_t(0) = 0, \gamma_t(1) = 1 \\
        \gamma_t(\tau_1) < \gamma_t(\tau_2), \enspace\text{if}\enspace \tau_1 < \tau_2
    \end{cases}
\end{equation}
The above conditions impose the boundary conditions, and imply that any $\gamma_t \in \Gamma$ is a monotonically increasing function. This property is also known as order-preserving which is important to time series alignment. It is easy to show that:
\begin{itemize}[noitemsep,nolistsep]
    \item[--] $\forall \gamma_{t_1}, \gamma_{t_2} \in \Gamma, \gamma_{t_1} \circ \gamma_{t_2} \in \Gamma$,
    \item[--] $\gamma_0 = \gamma_{id} \in \Gamma$,
    \item[--] $\forall \gamma_t \in \Gamma, \exists \inv{\gamma_t} \in \Gamma$ \textit{s.t.} $\gamma_t \circ \inv{\gamma_t} = \gamma_{id}$, where $\gamma_{id}(\tau) = \tau$ is an identity warping function. 
\end{itemize}
These properties imply that $\Gamma$ is a group under the operation of warping function composition. While in Dynamic Time Warping both constraints are satisfied as a monotonic path is computed from the initial to the endpoint, we have imposed these constraints to our ResNet-TW as will describe next. Inspired by~\cite{freifeld2017transformations} and on top of our Residual architecture, we define a finite tessellation, denoted by $\mathcal{T} = \{T_i\}_{i=1}^{N_\mathcal{T}}$ where $N_{\mathcal{T}} \in \mathbb{Z}_{+}$, to be a set of $N_{\mathcal{T}}$ closed subsets of $\Omega$, also named \textit{cells}. The union of $\mathcal{T}$ is $\Omega$ and the intersection of any adjacent cells is their shared border. For time-series, each cell is a 1-dimensional interval and has two vertices as borders. A velocity field $v_t$ for a given $t \in [0,1]$ is a map viewed as the mapping of points $\tau \in \Omega$ to $\tau^\prime \in \Omega$. An affine velocity field \textit{w.r.t.} $\mathcal{T}$ in cell $T_i$ is defined as $v_t^i(\tau) = A_t^i\tilde{\tau}$ where,
\begin{equation}
    \tilde{\tau} \triangleq 
    \begin{bmatrix}
        \tau \\
        1
    \end{bmatrix} \in \mathbb{R}^{n+1}, A_t^i \in \mathbb{R}^{n \times (n + 1)}\enspace. 
\end{equation}
For time series, $n=1$ and thus $A_t^i = [a_t^i, b_t^i]$. The velocity field $v_t$ is continuous if the continuity is ensured for points on borders, \ie, $A_t^i\tilde{\tau}_i = A_t^{i+1}\tilde{\tau}_i \implies a_t^i\tau_i + b_t^i = a_t^{i+1}\tau_i + b_t^{i+1}$ where $\tau_i$ is on the border of two pre-defined adjacent cells $T_i$ and $T_{i+1}$. All $\{b_t^i\}_{i=2}^{N_{\mathcal{T}}}$ can be computed through recursion by knowing $b_t^1$ and $\{a_t^i\}_{i=1}^{N_{\mathcal{T}}}$. Based on the above definitions, we design each \textit{projection} step as MLP which outputs $b_t^1$ and $\{a_t^i\}_{i=1}^{N_{\mathcal{T}}}$. (We discretize $t \in [0, 1]$ to correspond to $L$ blocks). The monotonic increasing (order-preserving) property of $\gamma_t$ can be assured by forcing each $a_t^i > 0$, which is achieved by using a ReLU or exponential operation before the output of each projection block. The boundary condition is realized by scaling $\gamma_t$ into the interval of $[0,1]$.


\section{Experiments}
\label{sec:exp}
In this section, we provide some experimental illustrations, validations, and relative evaluations of the proposed method on real datasets. We use a subset of 84 datasets from the UCR Archive ~\cite{chen2015ucr} for univariate time-series classification archive and three 3D Action/Activity recognition datasets – and Florence 3D \cite{seidenari2013recognizing}, MSR Action3D~\cite{li2010action} and MSR Daily Activity \cite{wang2012mining} for multivariate time-series classification.

\subsection{Univariate time-series (UCR archive datasets)}
\label{subsec:ucr}
The UCR time series classification archive contains 85 real-world datasets and we use a subset containing 84 datasets, as in \cite{weber2019diffeomorphic}. These datasets differ from each other in the number of examples, time-series length, application domain, (\eg, ECG, medical imaging, motion sensors), and number of classes ranging from 2 to 60. We experiment with the train and test split provided with the archive. Here we report a summary of our results which are fully detailed in our \textbf{supplementary materials}. For each of the UCR datasets, we train our ResNet-TW for joint alignment, where $\lambda_{var} \in \{10^{-3}, 10^{-2}\}$ and $\lambda_{smooth} \in \{0.5, 1\}$. Our ResNet-TW is composed of 4 to 8 building blocks for different datasets and each block consists of 3 convolutional layers with kernel size set to 51 and channel number set to 128. We set $\alpha=0.1$. The network is initialized by Xavier initialization using a normal distribution. We optimize our network with learning rate set to $10^{-4}$ and without weight decay. During the training, our ResNet-TW jointly aligns training samples for each class and computes a sample mean. In testing, we adopt the Nearest Centroid Classification (NCC) in which NCC is conducted by aligning first each test sample through the trained ResNet-TW (\ie our learned metric) and thus returns a distance to each of the training set centroids. 

\begin{wrapfigure}{r}{0.6\textwidth}
\begin{center}
\includegraphics[width=1\linewidth]{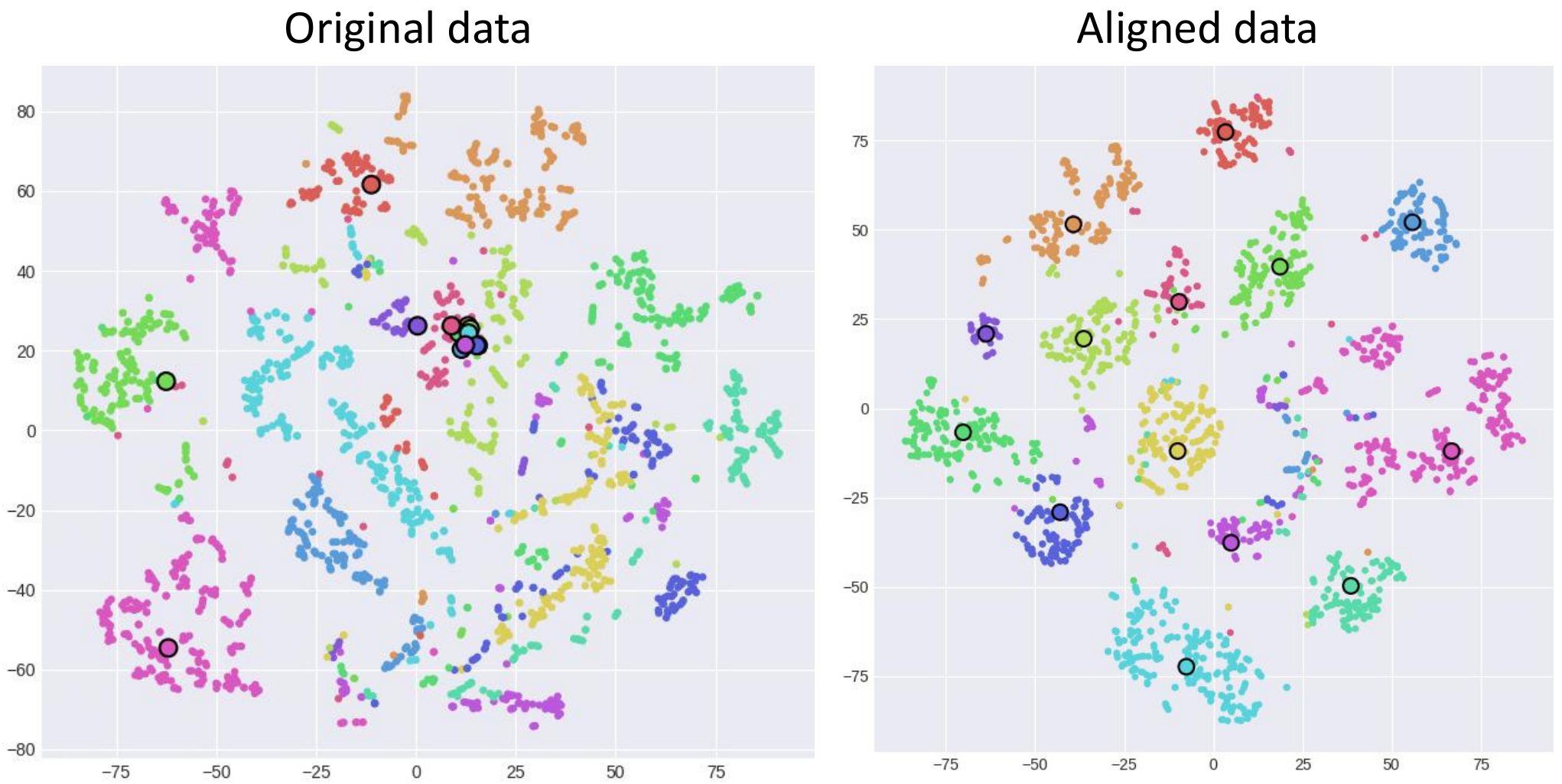}
\caption{t-SNE visualization of the original and aligned test data of the 11-class FacesUCR dataset. The dots with black borders are the \textit{average} of all samples within each class.}
\label{fig:ucr_tsne_facesucr}
\end{center}
\end{wrapfigure}

In Figure~\ref{fig:ucr_tsne_facesucr}, we provide a t-SNE visualization of the original and aligned data~\cite{van2008visualizing}, illustrates how our ResNet-TW decreases intra-class variance while increasing inter-class one, thus improving the performance of classification. We compare our ResNet-TW to: (1) the sample mean of the misaligned sets (Euclidean); (2) DBA~\cite{petitjean2011global}; (3) SoftDTW~\cite{cuturi2017soft} and (4) DTAN~\cite{weber2019diffeomorphic}. DBA and SoftDTW were measured by DTW distance, and DTAN is measured by Euclidean distance as ours. Figure~\ref{fig:ucr_acc} shows the NCC experimental results. Each point above the diagonal stands for an entire dataset of which our ResNet-TW correct classification rate is better than (or equal to) the competing method. This was the case for 89\% of the datasets when compared to Euclidean, 72\% for DBA, and 66\% for SoftDTW and 61\% for DTAN. These results (1) illustrate the importance of aligning the misaligned data for classification and (2) indicate that the average sequences of unwarped sequences synchronized by our ResNet-TW is usually more representative than other compared methods.

\begin{figure*}[!ht]
\begin{center}
\centerline{\includegraphics[width=1.0\linewidth]{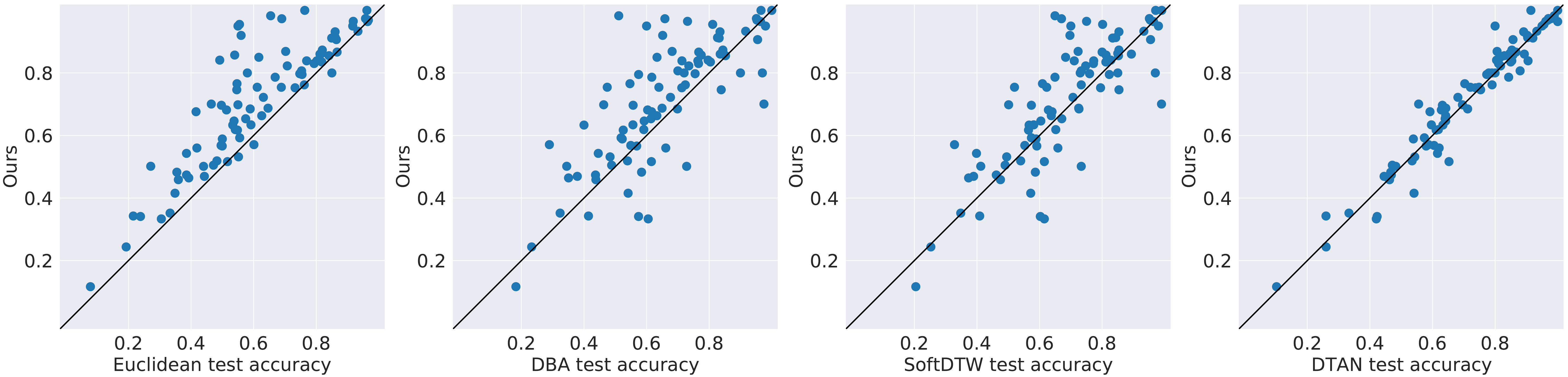}}
\caption{Correct classification rates using NCC. Each point above the diagonal indicates an entire UCR archive dataset where our ResNet-TW achieves better (or no worse) results than the comparing method. From left to right, our test accuracy compared with: Euclidean (ResNet-TW was better or no worse in 89\% of the datasets), DBA (72\%), SoftDTW (66\%) and DTAN (61\%).}
\label{fig:ucr_acc}
\end{center}
\vskip -0.2in
\end{figure*}

\subsection{Multivariate time-series (3D Action Recognition Datasets)}
\label{subsec:msr}

\textbf{Florence 3D-Actions dataset}~\cite{seidenari2013recognizing} is captured using a Kinect camera. It includes 9 activities: \textit{wave, drink from a bottle, answer phone, clap, tight lace, sit down, stand up, read watch, bow}. During acquisition, 10 subjects were asked to perform the above actions for two or three times. \textbf{MSR Action-3D dataset}~\cite{li2010action}  consists of a total of 20 types of segmented actions. It consists of a total of 20 types of segmented actions: \textit{high arm wave, horizontal arm wave, hammer, hand catch, forward punch, high throw, draw x, draw tick, draw circle, hand clap, two hand wave, side boxing, bend, forward kick, side kick, jogging, tennis swing, tennis serve, golf swing, pick up \& throw}. Each action starts and ends with a neutral pose and is performed two or three times by each of the 10 actors. \textbf{MSR Daily Activity dataset}~\cite{wang2012mining} is captured by a Kinect device and it consists of 16 activity types : \textit{drink, eat, read book, call cellphone, write on a paper, use laptop, use vacuum cleaner, cheer up, sit still, toss paper, play game, lay down on sofa, walk, play guitar, stand up, sit down}. The challenging part here is that each subject performs an activity in two different poses: sitting and standing. Following the cross-subject experimental setting, where the first five actors are taken as training and the last five for testing, we provide our quantitative results. Before, that we present some qualitative results. 

\head{Ablative study using synthetic warping.} For selected action sequences in Florence 3D, we randomly generate a synthetic warping function using \texttt{utility\_functions.rgam} function in fdasrsf\footnote{\url{https://fdasrsf-python.readthedocs.io}}, a python package for functional data analysis. We set the variance of warping functions to 10, such that we can synthesize warping functions with large temporal deformations. Next, we warp original action sequences $g$ using these synthetic warping functions as target sequences $f$. Finally, we adopt our ResNet-TW (for pairwise alignment) to estimate the synthetic warping functions $\gamma$ by minimizing the Euclidean distance between the warped original sequences $g \circ \gamma$ and target sequences $f$. 

\begin{figure}[!ht]
\includegraphics[width=.5\linewidth]{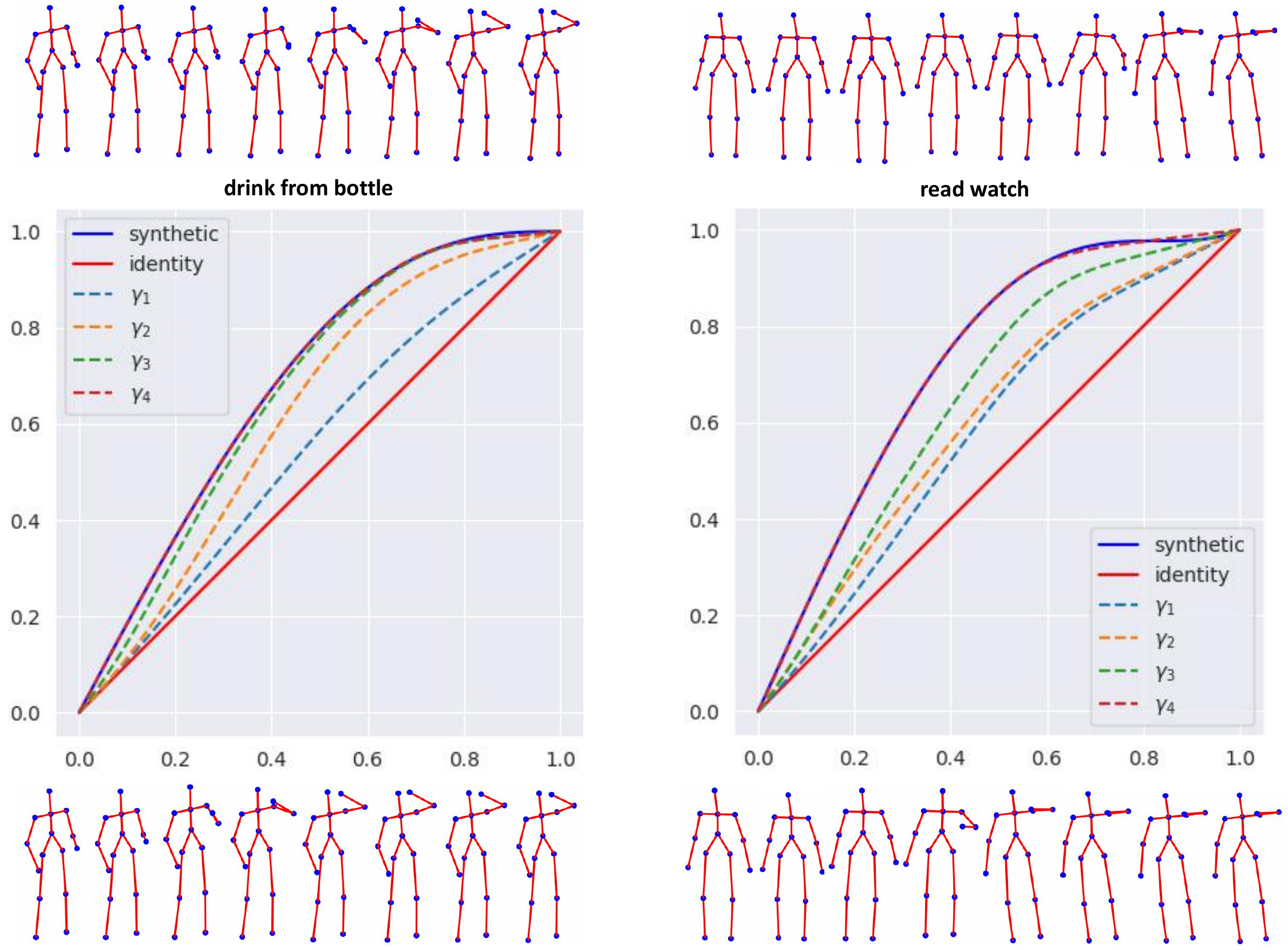}
\includegraphics[width=.5\linewidth]{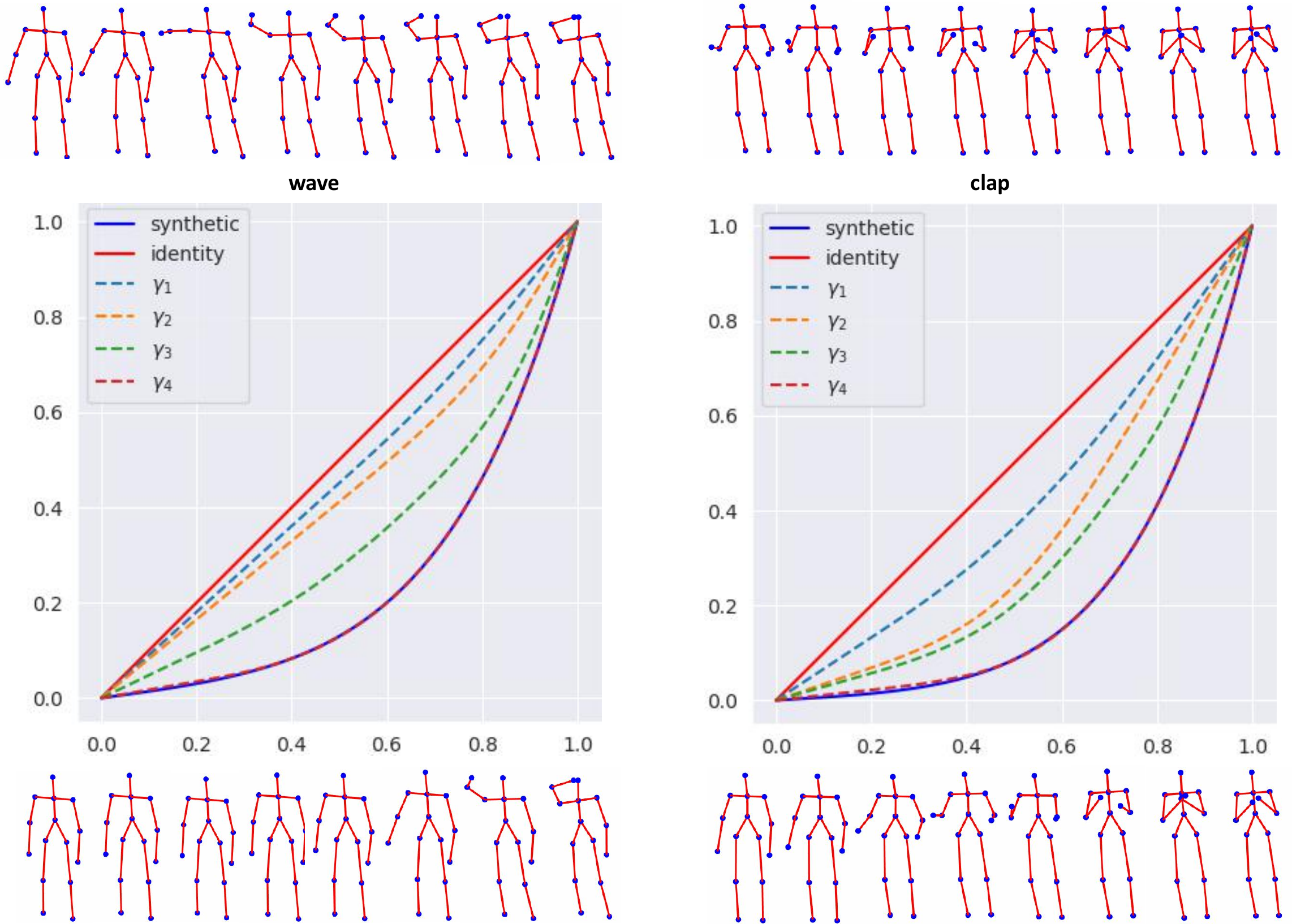}
\caption{Estimation of synthetic warping functions for examples from Florence 3D dataset. Top: Sampled frames from original action sequences. Middle: Estimated warping functions by each block of our ResNet-TW. Bottom: Corresponding sampled frames from synthetically reparameterized sequences.}
\label{fig:florence_syn}
\end{figure}

\begin{wrapfigure}{r}{0.5\textwidth}
\begin{center}
\includegraphics[width=\linewidth]{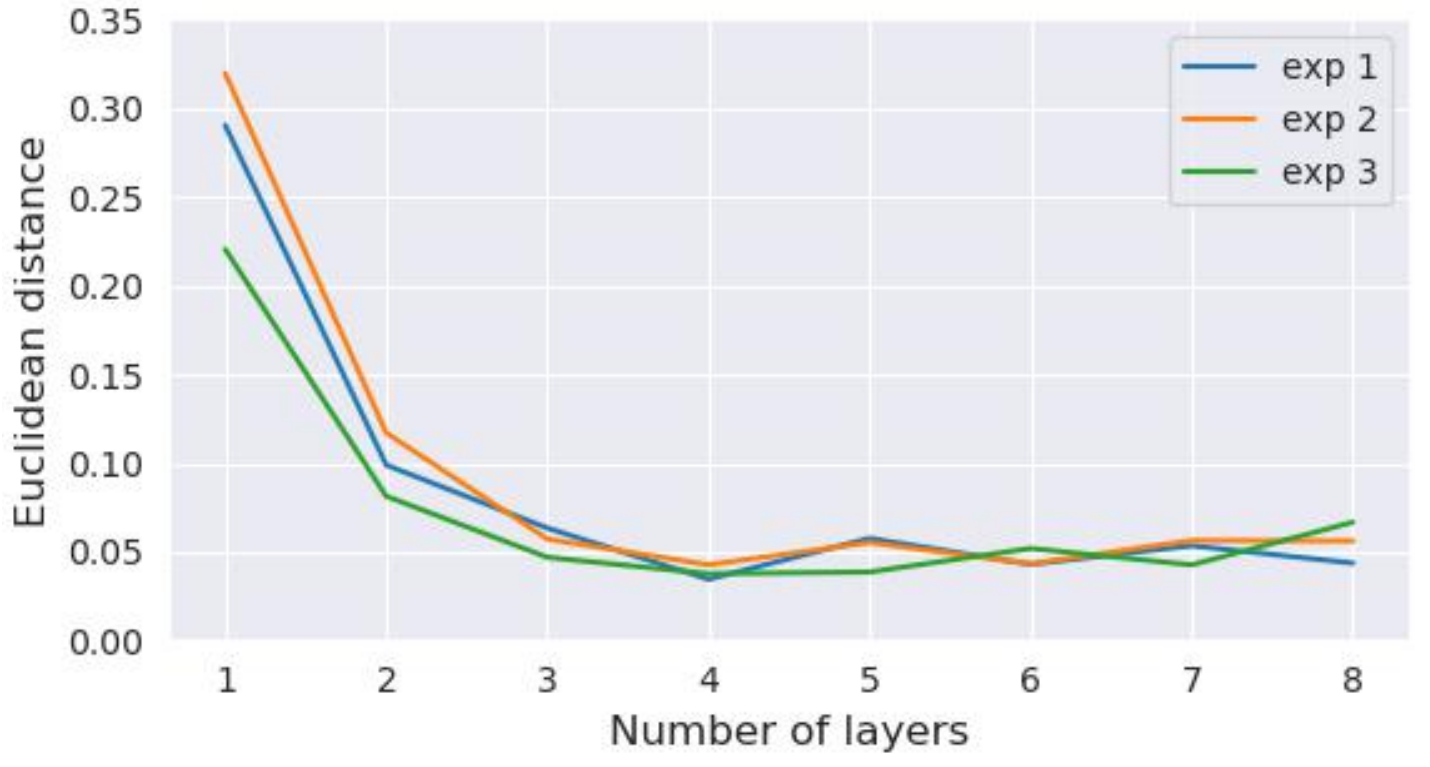}
\caption{Effect of the \# of building blocks in ResNet-TW on action sequences alignment.}
\label{fig:florence_layers}
\end{center}
\end{wrapfigure}

Four examples of estimated warping functions along with the ground-truth are shown in Figure~\ref{fig:florence_syn}. In particular, intermediate transformations $\gamma^l$, derived at each building block are shown. The overlaps of solid blue lines with red dashed lines show that our ResNet-TW is able to estimate (large) temporal deformations in an incremental fashion, \ie, the final warping function is sum of a serial of small velocity fields. We also notice that the smoothness of all dashed lines indicate each warping function is diffeomorphic, \ie, both the function and its inverse are differentiable.

We also evaluate the effect of the number of composed diffeomorphic mappings by varying the number of blocks in ResNet-TW from 1 to 8. We adopt the Euclidean distance between the target sequence $f$ and warped original sequence $g \circ \gamma^\ast$ as the evaluation metric. We run experiment for three times with different random seeds (to generate synthetic warping functions) and plot curves in Figure~\ref{fig:florence_layers}. We notice that with the number of blocks increasing, the distance dramatically decreases. However, when we use more than 4 blocks, the results are even slightly worse. It proves large temporal deformations can be approximated by a series of small deformations and excess compositions are less helpful.

\head{Quantitative evaluation.} Following the cross-subject experimental setting described in~\cite{li2010action} for MSR Action-3D dataset, we use the five first subjects for training and the last five for testing. All sequences are upsampled to 50 frames using the geometric toolbox proposed in~\cite{amor2015action}. The ResNet-TW is composed of 8 building blocks and each block consists of 3 convolutional layers with kernel size set to 51 and channel number set to 128. We set $\alpha=10^{-4}, \lambda_{\textrm{smooth}}=1$ and $\lambda_{\textrm{var}}=0.05$. The network is initialized by Xavier initialization using a normal distribution. We optimize our network with learning rate set to $10^{-5}$ and without weight decay. We report in the following our quantitative results on the three action recognition datasets  accompanied with a comprehensive comparative study w.r.t recent approaches, in particular \cite{weber2019diffeomorphic} and \cite{su2020learning}. To allow fair comparison, we follow the experimental settings of \cite{li2010action},  \cite{seidenari2013recognizing} and \cite{wang2012mining}, respectively. We have also conducted similar experiments using DTAN/R-DTAN proposed in \cite{weber2019diffeomorphic} and RVSML-OPW/RVSML-DTW introduced in \cite{su2020learning} on all datasets. 

\begin{table}[ht!]
\caption{Comparison w.r.t. recent approaches on Florence 3D - MRS Action3D - Daily Activity 3D.}
\begin{footnotesize}
\begin{tabular}{|l|c|c|c|c|c|}
\hline
\textbf{Method/Metric} &  Year & \textbf{NCC (NM)} & \textbf{1-NN} & \textbf{3-NN} & \textbf{5-NN} \\
\hline
\hline
DTAN \cite{weber2019diffeomorphic} & 2019 & 0.60 - 0.71 - 0.49 &  \textbf{0.76} - 0.72 - 0.58 & 0.72 - 0.69 - 0.51 & 0.69 - 0.65 - 0.53   \\
R-DTAN \cite{weber2019diffeomorphic} & 2019 &  0.63 - 0.71 - 0.50 &  \textbf{0.76} - 0.72 - \textbf{0.61} & \textbf{0.73} - 0.69 - \textbf{0.54} & \textbf{0.71} - 0.66 - 0.53  \\
\hline
\hline
RVSML-OPW \cite{su2020learning} & 2020  &  0.57 - 0.63 - 0.43 &  0.75 - 0.68 - 0.58 & 0.70 - 0.69 - 0.52 & 0.68 - 0.62 - 0.51   \\
RVSML-DTW \cite{su2020learning} & 2020 &  0.59 - 0.56 - \textbf{0.52} &  0.66 - 0.74 - 0.58 & 0.68 - \textbf{0.77} - 0.51 & 0.60 - \textbf{0.73} - 0.48\\
\hline
\hline
\textbf{ResNet-TW} & -- & \textbf{0.64} - \textbf{0.75} - 0.50 & \textbf{0.76} - \textbf{0.76} - \textbf{0.61} & 0.72 - 0.70 - \textbf{0.54} & \textbf{0.71} - 0.71 - \textbf{0.55} \\
\hline
\end{tabular}
\end{footnotesize}
\label{tab:Comparison}
\end{table}

In table.~\ref{tab:Comparison}, we report the classification accuracy of the proposed ResNet-TW associated with different classifiers (NCC/NM: Nearest Mean), $k$-NN ($k$-Nearest Neighbor, with $k\in \{1,3,5\}$). We have ran all methods on the same data features and use exactly same experimental settings. These comparative studies show the competitiveness of our ResNet-TW, in particular with Nearest Mean and $1$-NN classifiers with 4\% improvement seen on MSR Action 3D. While DTAN consists of three steps (i) a Localization Network, (ii) a Grid generator and (iii) a differentiable time-series resampler, ResNet-TW is much simpler and consists of identical building blocks (Deep ReLu NNs) optimized within a same and unique residual architecture. While R-DTAN computes stationary velocity fields as the same network is trained (i.e. the approach is static just like log-Euclidean polyaffine approach proposed in \cite{arsigny2006log}, ResNet-TW is dynamic and compute non-stationary velocity fields. This allows more flexibility and more accuracy especially in case of large temporal variations (\cite{younes2010shapes}, p. 292). Importantly, unlike DTAN and R-DTAN, our ResNet-TW minimizes kinetic energy as a part of its regularizer (Eq.\ref{eq:lddmm_v2gamma}). Thus, not only plausible transformations but also optimal (in terms of energy) warping functions are generated. As demonstrated in \cite{younes2010shapes}, Theorem 8.7) the kinetic energy (length of the path connecting the identity to $\gamma$) is a proper metric and provides a measure of distance for the space of diffeomorphisms Diff$_V$, while $V$ is an admissible space velocity fields. This is another advantage of ResNet-TW compared to R-DTAN.

\section{Conclusion}
\label{sec:Conclusion}
We proposed a diffeomorphic temporal transformer for both pairwise (unsupervised) and joint (semi-supervised) sequences alignment. Our ResNet-TW estimates time-warping functions through an integration of smooth and regular velocity fields, building blocks of the residual architecture, and thus incrementally computes the final warping. Geometrically, our ResNet-TW is an Eulerian discretization of the ODE (\ie the flow equation) which governs the final reparameterization (temporal warping) transformation. Regularized neural functions guarantees smooth and regular velocity fields. Intermediate warping functions are also diffeomorphic. Experiments on several datasets validates our ResNet-TW. It does not only align pairwise sequential data but also is capable of learning representative \textit{sample average} sequences for multi-class joint alignment. Our ResNet-TW builds a step further in bridging between Deep Residual Networks and geometric diffeomorphic frameworks (\ie, LDDMM \cite{beg2005computing}). We leave the comparison of the static (Gaussian) kernel used in LDDMM and the dynamic kernels inferred by the building blocks of our ResNet and induced RKHS \cite{bietti2017invariance} to guarantee smooth and regular velocity fields \cite{miller2006geodesic} for a future investigation. %

\bibliographystyle{plain}
\bibliography{neurips_2021}

\clearpage

\appendix

\section{Additional Qualitative Results on MSR Action-3D}
In this section, we provide more qualitative results on MSR Action-3D dataset for both pairwise and multi-class joint alignment.

\subsection{Pairwise Alignment}
For each action category, we randomly pick two sequences performed by different actors, thus, there exists temporal misalignment between these two sequences. We treat one sequence as a query $g$ and another as
a target $f$. We adopt our ResNet-TW to estimate the warping function by minimizing the Euclidean distance between the warped query sequence $g \circ \gamma$ and target sequence $f$. Figure~\ref{fig:msr_pair} illustrates three examples of pairwise alignment. 

\subsection{Joint Alignment}
\begin{figure}[!ht]
    \centering
    \begin{minipage}{0.52\textwidth}
        \centering
        \includegraphics[width=0.9\textwidth]{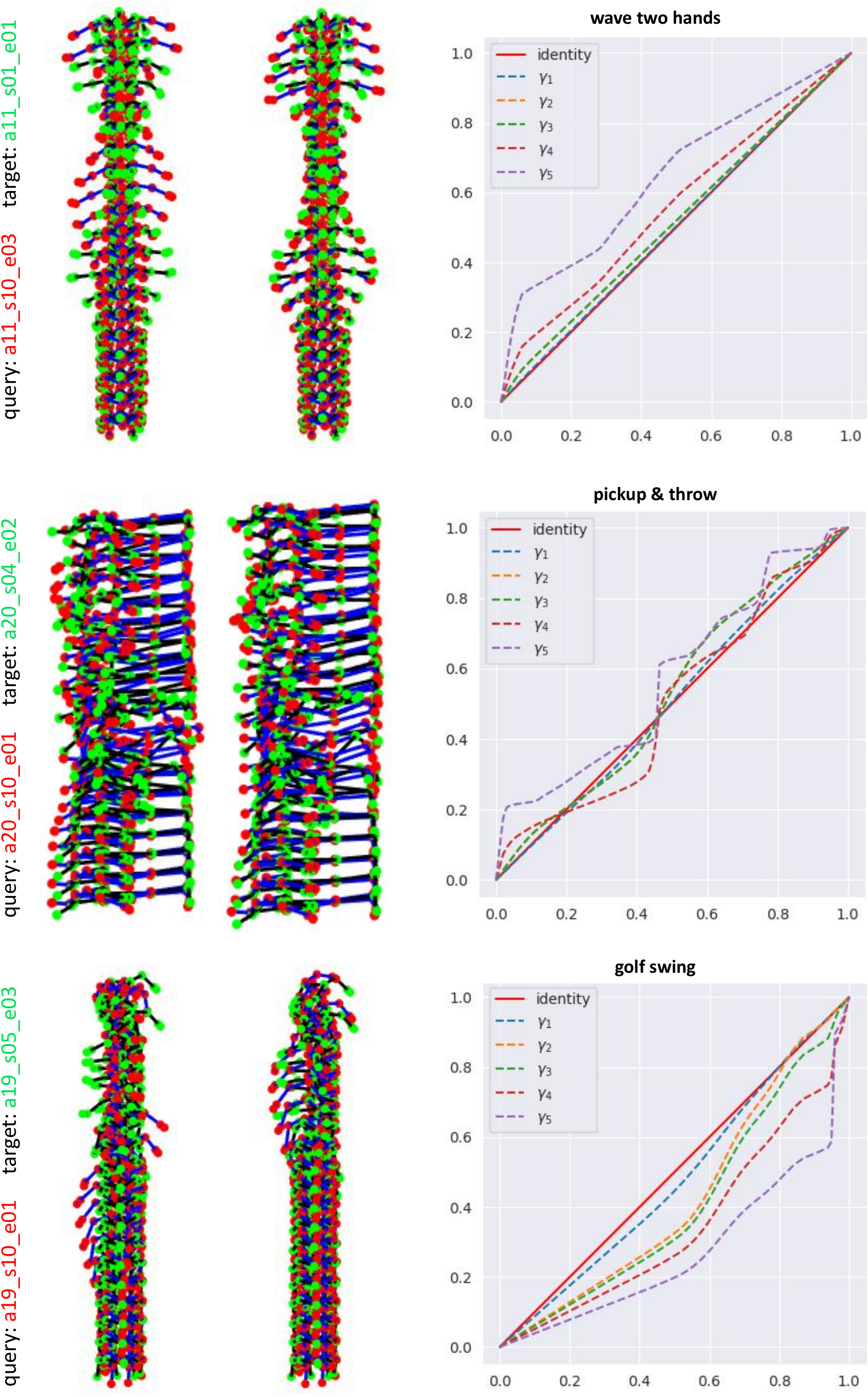}
        \caption{Results of pairwise alignment on MSR Action-3D dataset. In each case, $f$ and $g$ (first column), $f$ and $g \circ \gamma^\ast$ (second column). We note that the temporal misalignment of the red and the green skeleton sequences is clear in the left column. In contrast, they are well aligned in the second column after temporal registration. The plot in the third column shows the optimal warping functions $\gamma^\ast$ obtained using our ResNet-TW.}
        \label{fig:msr_pair}
    \end{minipage}\hfill
    \begin{minipage}{0.45\textwidth}
    \begin{center}
    \subfloat[Hammer; top: a03\_s01\_e01; middle: a03\_s06\_e01.]{
	\label{fig:msr_hammer}
	\centerline{\includegraphics[width=1.0\linewidth]{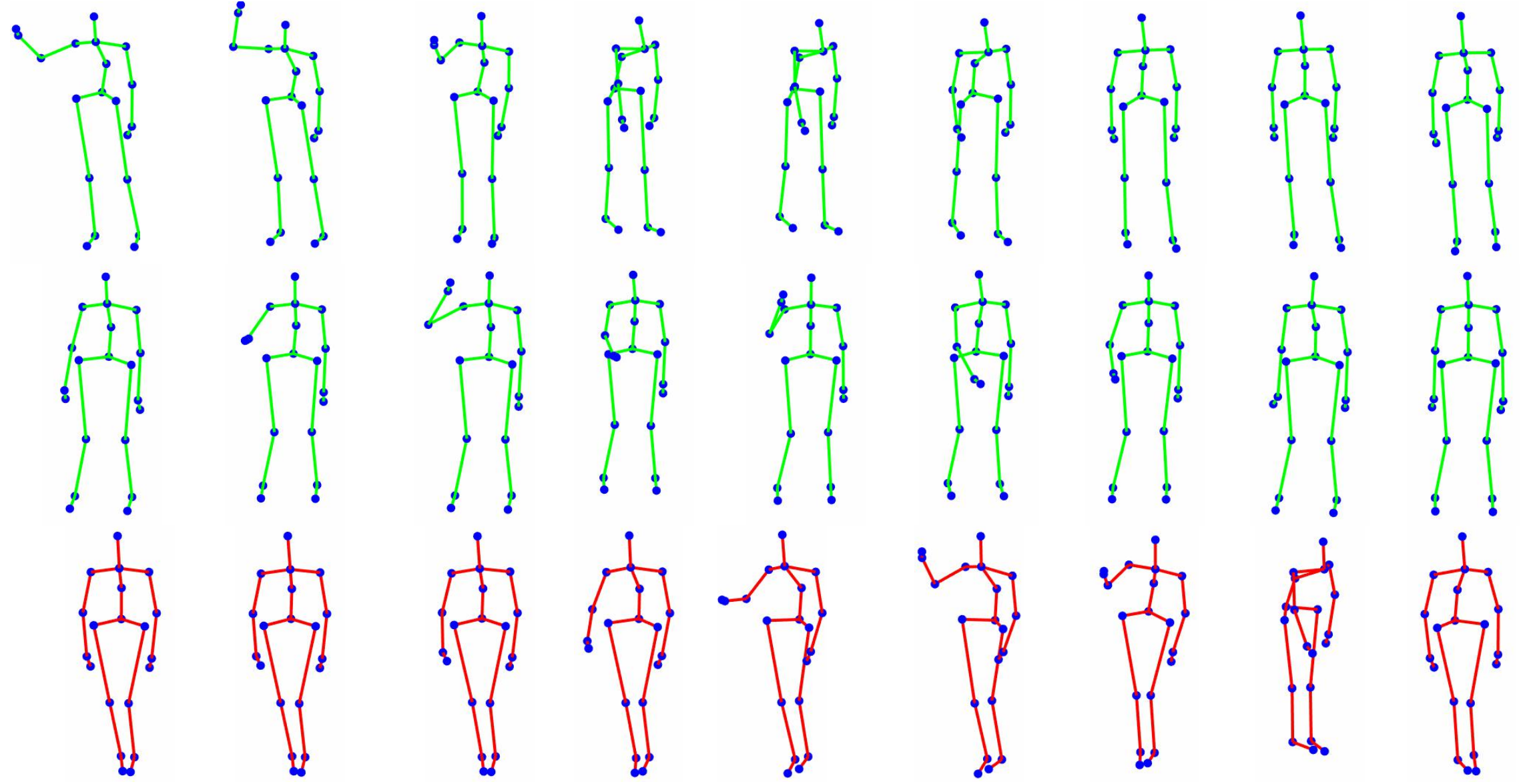}}}
	
    \subfloat[High arm wave; top: a01\_s05\_e02; middle: a01\_s07\_e01.]{
	\label{fig:msr_high_arm_wave}
	\centerline{\includegraphics[width=1.0\linewidth]{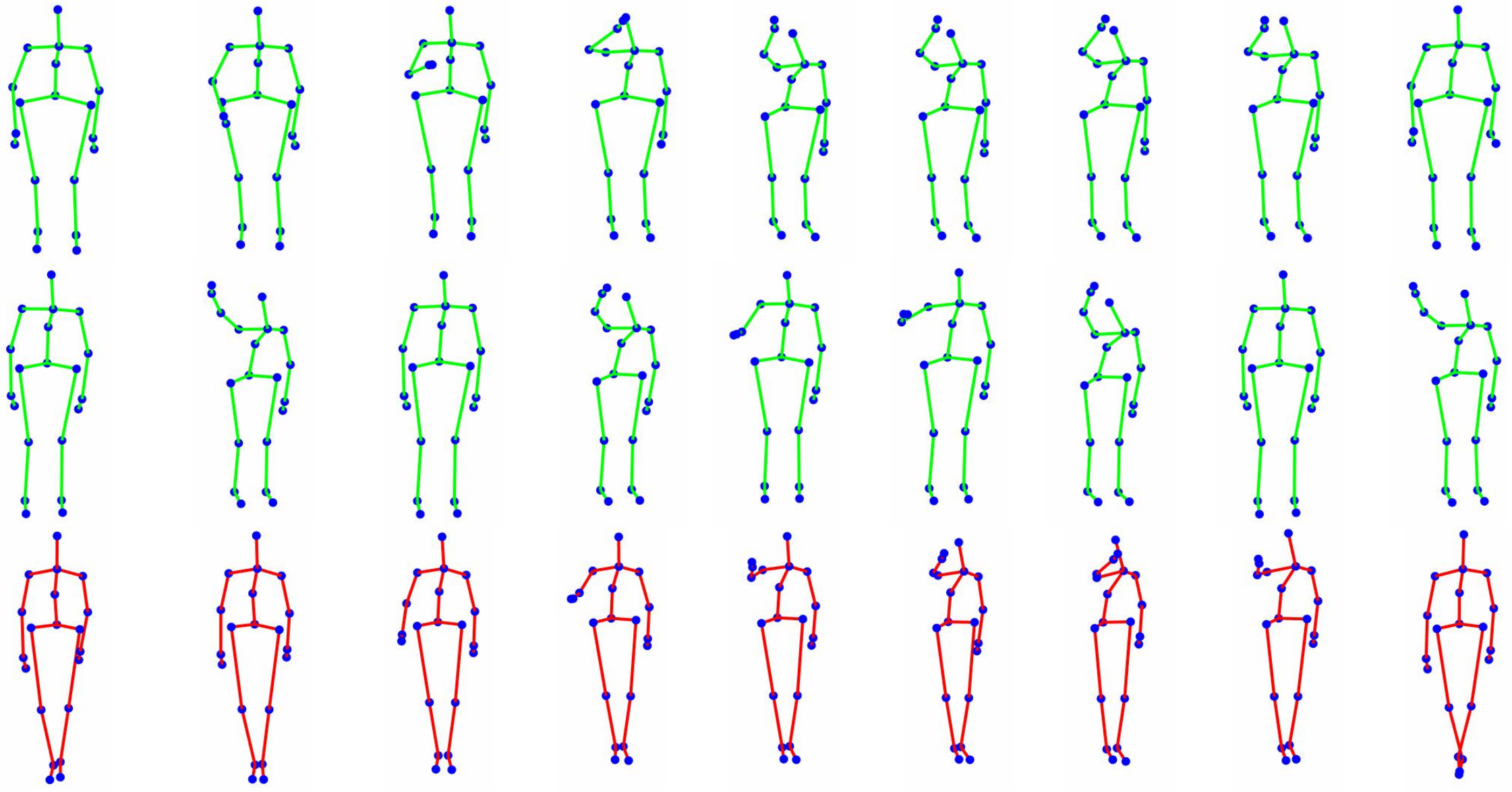}}}
	
    \subfloat[Side tick; top: a15\_s07\_e03; middle: a15\_s09\_e01.]{
	\label{fig:msr_side_kick}
	\centerline{\includegraphics[width=1.0\linewidth]{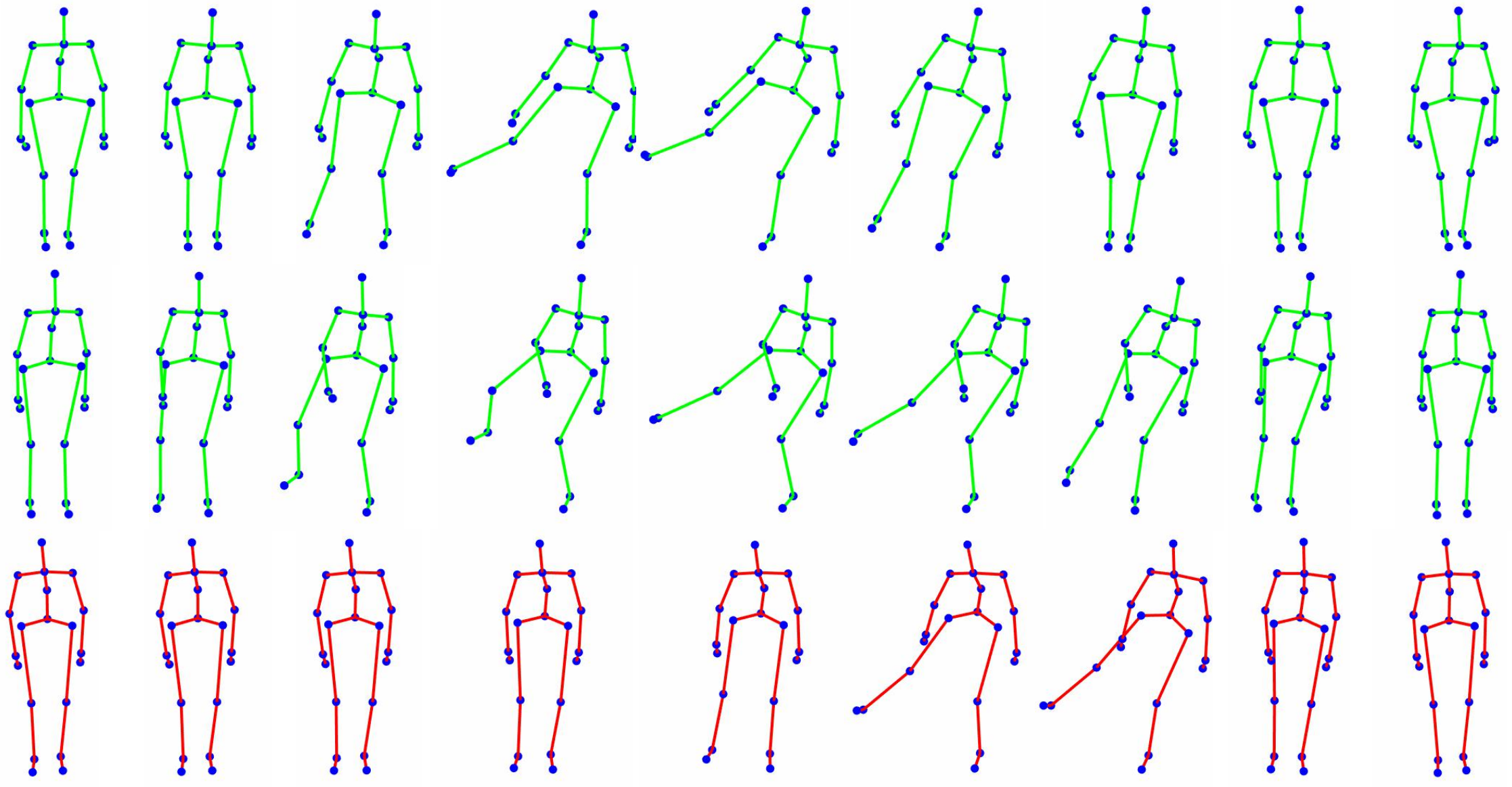}}}
	
    \caption{Joint multi-class alignment of MSR Action dataset. Green: original action sequences. Red: averaged sequences.}
    \label{fig:msr_joint_align}
    \end{center}
    \end{minipage}
\end{figure}

In this subsection, we investigate whether our ResNet-TW can learn the ``average'' action sequences within categories. We use the five first subjects for training and the last five for testing. All sequences are up-sampled to 50 frames using the pipeline proposed in~\cite{amor2015action}. Six action sequences (green) from three different categories are sampled and visualized in Figure~\ref{fig:msr_joint_align}. We also visualize the average action sequences (red). Note
that the average sequences are averaged over all testing
samples in the corresponding categories, not the sampled
two sequences. From Figure~\ref{fig:msr_joint_align}, we notice that the average sequences capture key poses of each action, showing the ability of our ResNet-TW to capture essential patterns of each action. Although the movement amplitudes of the average sequences decrease, we attribute such phenomena to the counteraction of spatial variability of actions performed by different actors.

\clearpage
\section{Additional Alignment Results of Test Data}
In this section, we provide more qualitative results of joint alignment of test data in different datasets from the UCR archive~\cite{chen2015ucr}.

\subsection{Within-class Variance Reduction}
\begin{figure}[!hb]
    \centering
    \begin{subfigure}{0.95\textwidth}
        \includegraphics[width=1.0\linewidth]{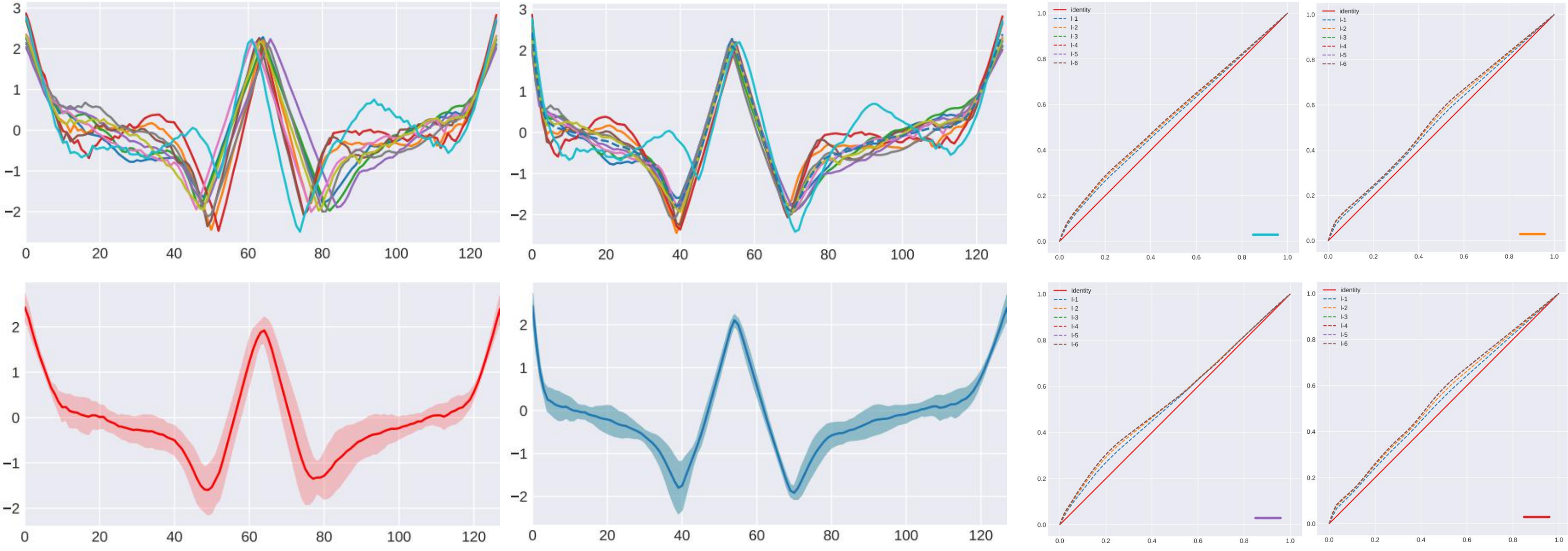}
        \subcaption{SwedishLeaf}
        \label{fig:ucr_swedishleaf}
    \end{subfigure}

\medskip
    \begin{subfigure}{0.95\textwidth}
        \includegraphics[width=1.0\linewidth]{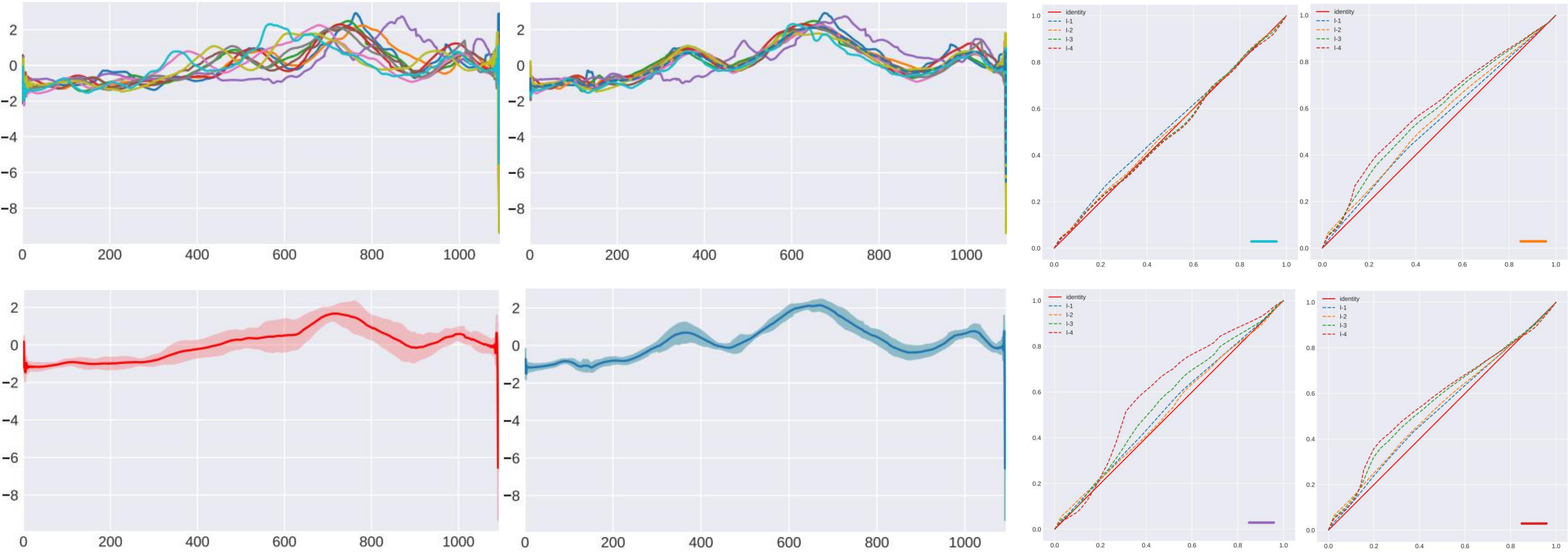}
        \subcaption{Haptic}
        \label{fig:ucr_haptic}
    \end{subfigure}
    \caption{Within-class joint alignment of four datasets from the UCR archive~\cite{chen2015ucr}. All results are randomly selected from previously unseen test data. \textbf{First column}: misaligned signals (top), the average signal and ± standard deviation shown in red shaded area (bottom). \textbf{Second column}: aligned signals (top), the average signal and ± standard deviation shown in blue shaded area (bottom). \textbf{Third and fourth columns}: warping functions $\gamma$ produced by each layer of our ResNet-TW. The colors of short bars at the right-bottom on the third and fourth columns correspond to signals on the first and second columns.}
\end{figure}%

\begin{figure}[ht]\ContinuedFloat
    \centering
    \begin{subfigure}{0.95\textwidth}
        \includegraphics[width=1.0\linewidth]{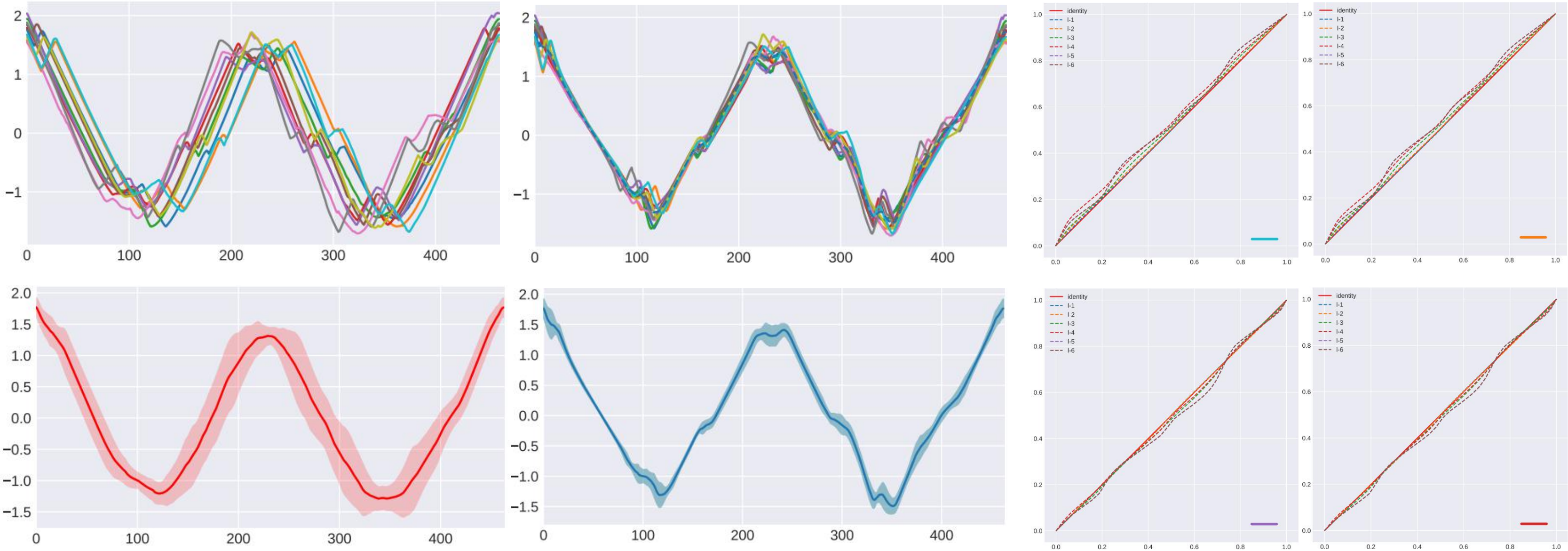}
        \subcaption{FISH}
        \label{fig:ucr_fish}
    \end{subfigure}

\medskip
    \begin{subfigure}{0.95\textwidth}
        \includegraphics[width=1.0\linewidth]{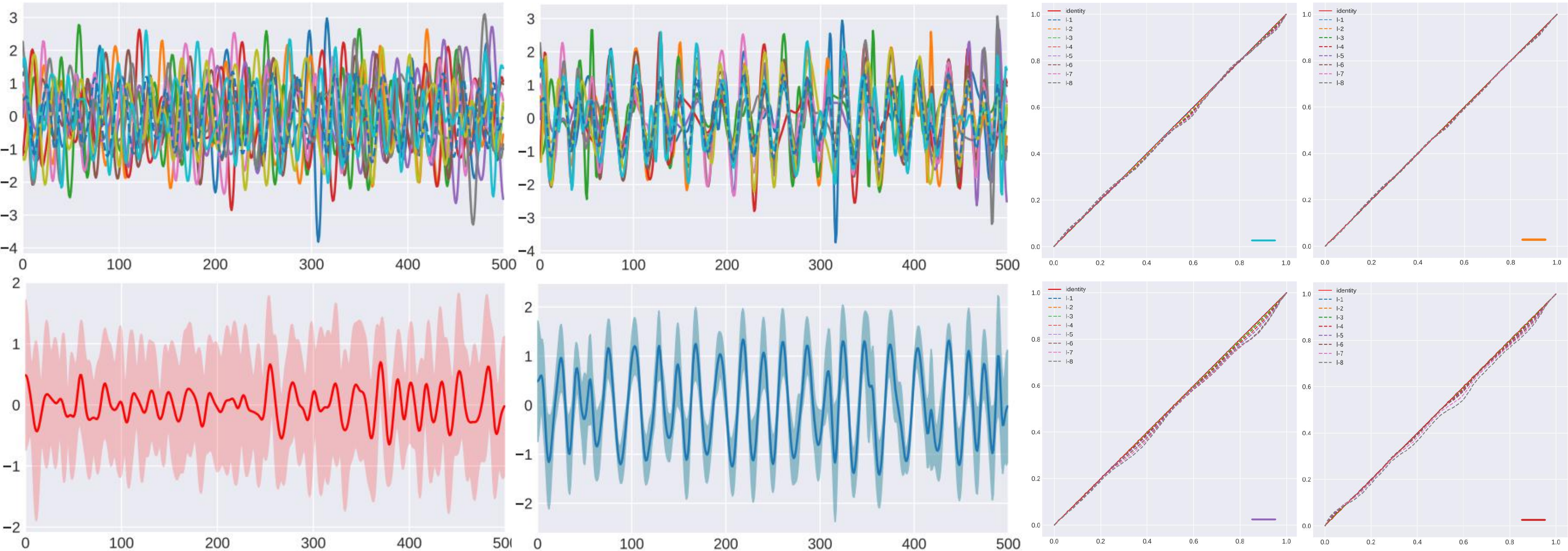}
        \subcaption{FordA}
        \label{fig:ucr_fordA}
    \end{subfigure}
    \caption[]{Within-class joint alignment of four datasets from the UCR archive~\cite{chen2015ucr}. All results are randomly selected from previously unseen test data. \textbf{First column}: misaligned signals (top), the average signal and ± standard deviation shown in red shaded area (bottom). \textbf{Second column}: aligned signals (top), the average signal and ± standard deviation shown in blue shaded area (bottom). \textbf{Third and fourth columns}: warping functions $\gamma$ produced by each layer of our ResNet-TW. The colors of short bars at the right-bottom on the third and fourth columns correspond to signals on the first and second columns.}
    \label{fig:arms}
\end{figure}
\clearpage

\subsection{Aligned Signals at Different Blocks in ResNet-TW}
\begin{figure}[!hb]  
\centering
\includegraphics[width=0.82\linewidth]{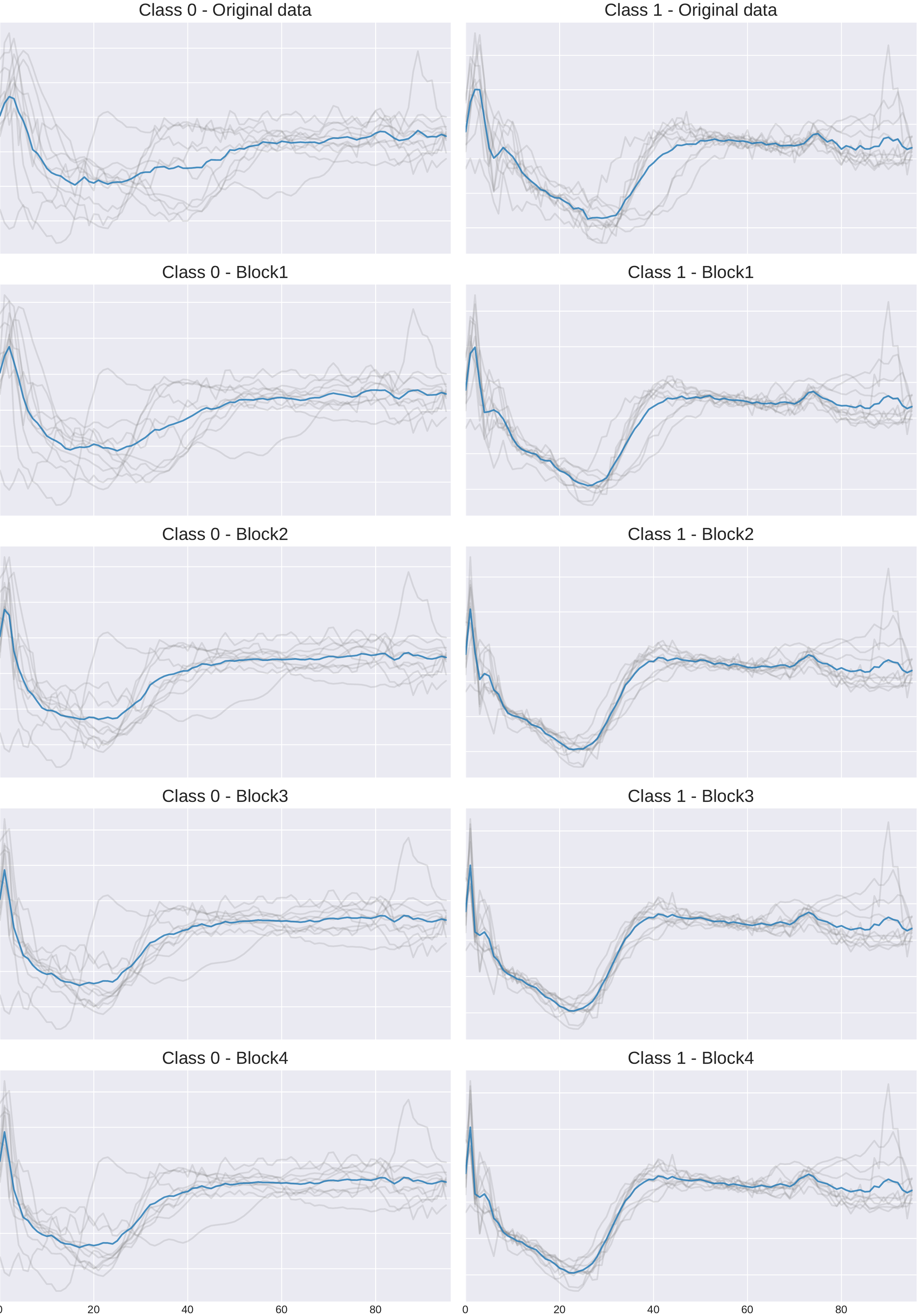}
    \caption{Examples of ECG200 dataset from the UCR archive~\cite{chen2015ucr}. First row: 10 randomly selected test data from each class of the dataset. Second until the last rows: aligned signals using the warping functions $\gamma$ produece by each block of our ResNet-TW. The blue curve represents the sample mean of the signals.}
    \label{fig:ucr_ecg200}
\end{figure}
%

\section{Nearest Centroid Classification (NCC) Results}
In this section, we show detailed quantitative results of the NCC experiment on UCR archive~\cite{chen2015ucr} in Figure~\ref{fig:ucr_acc} and Table~\ref{tab:ucr_result}\footnote{We update our results in this document.}. For the baseline experiment we used the Euclidean mean of the misaligned set. We compare our ResNet-TW with DTW Barycenter Averaging (DBA)~\cite{petitjean2011global}, SoftDTW~\cite{cuturi2017soft}, DTAN~\cite{weber2019diffeomorphic}.

\begin{figure}[!ht]
\begin{center}
\centerline{\includegraphics[width=1.0\linewidth]{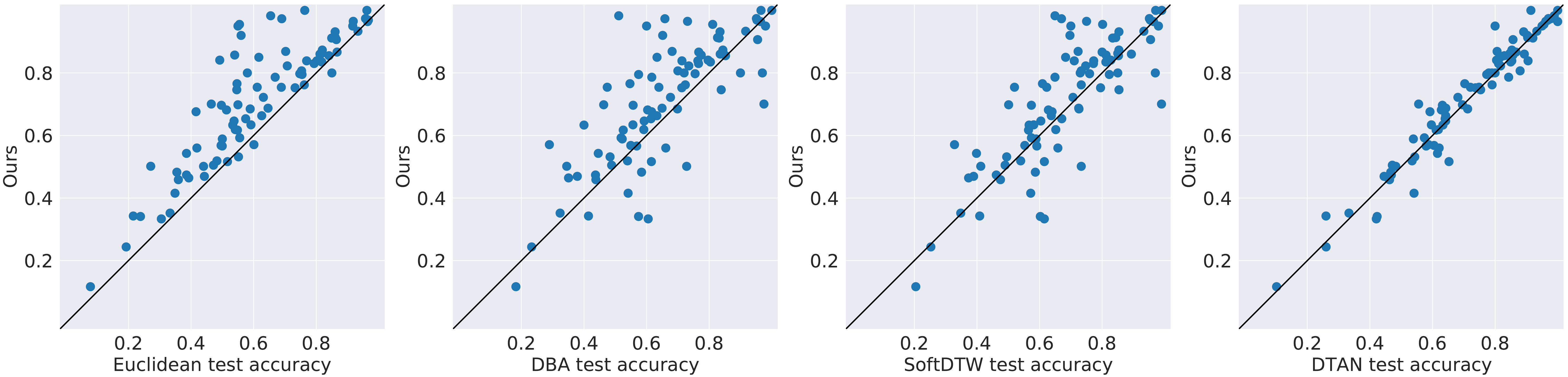}}
\caption{Correct classification rates using NCC on UCR archive~\cite{chen2015ucr} Each point above the diagonal indicates an entire UCR archive dataset where our ResNet-TW achieves better (or no worse) results than the comparing method. From left to right, our test accuracy compared with: Euclidean (ResNet-TW was better or no worse in 96\% of the datasets), DBA (79\%), SoftDTW (69\%) and DTAN (68\%).}
\label{fig:ucr_acc}
\end{center}
\vskip -0.2in
\end{figure}
%

%
\LTcapwidth=\textwidth
\begin{longtable}{lllllc}
\hline\hline
\textbf{Dataset}               & \textbf{Baseline} & \textbf{Softdtw} & \textbf{DBA} & \textbf{DTAN} & \multicolumn{1}{l}{\textbf{Ours}} \\ \hline \endhead 
50words                        & 0.516484          & 0.615385         & 0.615385     & \textbf{0.652747}& 0.516484                          \\
Adiac                          & 0.549872          & 0.501279         & 0.462916     & 0.695652      & \textbf{0.698210}                           \\
ArrowHead                      & 0.611429          & 0.520000         & 0.474286     & 0.748571      & \textbf{0.754286}                          \\
Beef                           & 0.533333          & 0.566667         & 0.400000     & 0.633333      & \textbf{0.633333}                          \\
BeetleFly                      & 0.850000          & 0.850000         & \textbf{0.900000} & 0.800000 & 0.800000                               \\
BirdChicken                    & 0.550000          & 0.700000         & 0.600000     & 0.800000      & \textbf{0.950000}                              \\
CBF                            & 0.763333          & \textbf{0.971111}& 0.965556     & 0.914444      & 0.850000                              \\
Car                            & 0.616667          & 0.683333         & 0.633333     & 0.816667      & \textbf{1.000000}                            \\
ChlorineConcentration          & 0.333073          & 0.348177         & 0.323698     & 0.333073      & \textbf{0.351823}                          \\
CinC\_ECG\_torso               & 0.385507          & 0.398551         & 0.445652     & \textbf{0.615942}      & 0.542754                          \\
Coffee                         & 0.964286          & 0.964286         & 0.964286     & \textbf{1.000000}      & 0.964290              \\
Computers                      & 0.416000          & 0.640000         & 0.616000     & 0.592000      & \textbf{0.676000}                             \\
Cricket\_X                     & 0.238462          & \textbf{0.602564}         & 0.574359     & 0.423077      & 0.341026                          \\
Cricket\_Y                     & 0.348718          & \textbf{0.571795}         & 0.541026     & 0.541026      & 0.415385                          \\
Cricket\_Z                     & 0.305128          & \textbf{0.615385}         & 0.605128     & 0.420513      & 0.333333                          \\
DiatomSizeReduction            & 0.957516          & 0.950980         & 0.950980     & 0.970588      & \textbf{0.973856}                          \\
DistalPhalanxOutlineAgeGroup   & 0.817500          & 0.850000         & 0.840000     & 0.847500      & \textbf{0.862500}                            \\
DistalPhalanxOutlineCorrect    & 0.471667          & 0.490000         & 0.488333     & 0.471667      & \textbf{0.505000}                             \\
DistalPhalanxTW                & 0.747500          & 0.760000         & 0.755000     & 0.780000      & \textbf{0.797500}                            \\
ECG200                         & 0.750000          & 0.730000         & 0.720000     & 0.790000      & \textbf{0.795031}                          \\
ECG5000                        & 0.860444          & 0.853778         & 0.834667     & \textbf{0.891333}      & 0.800000                               \\
ECGFiveDays                    & 0.689895          & 0.670151         & 0.658537     & \textbf{0.977933}      & 0.931556                          \\
Earthquakes                    & 0.754658          & 0.822981         & 0.574534     & 0.773292      & \textbf{0.973287}                          \\
ElectricDevices                & 0.482687          & 0.539748         & 0.538970     & \textbf{0.534820}      & 0.518869                               \\
FaceAll                        & 0.491716          & 0.827811         & 0.796450     & 0.804734      & \textbf{0.840909}                          \\
FaceFour                       & 0.840909          & 0.852273         & 0.852273     & 0.829545      & \textbf{0.855122}                          \\
FacesUCR                       & 0.539512          & 0.812683         & 0.774634     & 0.857073      & \textbf{0.857143}                         \\
FISH                           & 0.560000          & 0.697143         & 0.651429     & 0.902857      & \textbf{0.902857}                          \\
FordA                          & 0.495973          & 0.552902         & 0.549570     & \textbf{0.604832}      & 0.568176                          \\
FordB                          & 0.499725          & \textbf{0.591309}& 0.568482     & 0.579758      & 0.566282                          \\
Gun\_Point                     & 0.753333          & 0.733333         & 0.700000     & \textbf{0.880000} & 0.806667                          \\
Ham                            & 0.761905          & 0.733333         & 0.723810     & \textbf{0.790476}      & 0.761905                          \\
HandOutlines                   & 0.818000          & 0.812000         & 0.804000     & \textbf{0.850000}      & 0.835000                             \\
Haptics                        & 0.392857          & 0.373377         & 0.350649     & 0.457792      & \textbf{0.464286}                          \\
Herring                        & 0.546875          & 0.609375         & 0.546875     & 0.703125      & \textbf{0.765625}                          \\
InlineSkate                    & 0.192727          & 0.252727         & 0.232727     & \textbf{0.260000}      & 0.243636                          \\
InsectWingbeatSound            & \textbf{0.601010}          & 0.328283         & 0.289394     & 0.587374      & 0.570707                          \\
ItalyPowerDemand               & 0.918367          & 0.750243         & 0.730807     & 0.962099      & \textbf{0.965015}                                  \\
LargeKitchenAppliances         & 0.440000          & \textbf{0.733333}         & 0.728000     & 0.482667      & 0.501333                          \\
Lighting2                      & 0.688525          & 0.622951         & 0.639344     & 0.721311      & \textbf{0.754098}                          \\
Lighting7                      & 0.589041          & \textbf{0.726027}         & 0.698630     & 0.712329      & 0.684932                          \\
MALLAT                         & 0.966738          & 0.953945         & 0.952665     & \textbf{0.968870}      & 0.966738                          \\
Meat                           & 0.933333          & 0.933333         & 0.916667     & 0.933333      & \textbf{0.933333}                          \\
MedicalImages                  & 0.385526          & 0.461842         & 0.436842     & 0.468421      & \textbf{0.473684}                          \\
MiddlePhalanxOutlineAgeGroup   & 0.732500          & \textbf{0.795000}& 0.712500     & 0.737500      & 0.752500                            \\
MiddlePhalanxOutlineCorrect    & \textbf{0.551667} & 0.495000         & 0.483333     & 0.543333      & 0.531667                          \\
MiddlePhalanxTW                & 0.591479          & 0.581454         & 0.556391     & 0.596491      & \textbf{0.634085}                          \\
MoteStrain                     & 0.861022          & 0.843450         & 0.826677     & 0.904153      & \textbf{0.912939}                           \\
NonInvasiveFatalECGThorax1     & 0.769466          & 0.710941         & 0.712977     & \textbf{0.853435}      & 0.838677                          \\
NonInvasiveFatalECGThorax2     & 0.802036          & 0.773028         & 0.763868     & \textbf{0.905344}      & 0.838680                               \\
OliveOil                       & 0.866667          & 0.800000         & 0.766667     & 0.866667      & \textbf{0.866667}                          \\
OSULeaf                        & 0.359504          & 0.475207         & 0.438017     & \textbf{0.462810}      & 0.458678                          \\
PhalangesOutlinesCorrect       & 0.625874          & 0.637529         & 0.632867     & 0.642191      & \textbf{0.663170}                           \\
Phoneme                        & 0.078586          & \textbf{0.204641}& 0.182489     & 0.101793      & 0.116561                          \\
Plane                          & 0.961905          & 0.990476         & 1.000000     & 1.000000      & \textbf{1.000000}                          \\
ProximalPhalanxOutlineAgeGroup & 0.819512          & 0.853659         & 0.843902     & 0.853659      & \textbf{0.873171}                          \\
ProximalPhalanxOutlineCorrect  & 0.646048          & 0.725086         & 0.649485     & 0.642612      & \textbf{0.687285}                          \\
ProximalPhalanxTW              & 0.707500          & 0.747500         & 0.735000     & 0.817500      & \textbf{0.822500}                            \\
RefrigerationDevices           & 0.354667          & \textbf{0.586667}& 0.584000     & 0.466667      & 0.482667                          \\
ScreenType                     & 0.442667          & 0.389333         & 0.378667     & 0.445333      & \textbf{0.469333}                          \\
ShapeletSim                    & 0.500000          & \textbf{0.588889}& 0.522222     & 0.538889      & \textbf{0.588889}                          \\
ShapesAll                      & 0.513333          & 0.628333         & 0.603333     & 0.628333      & \textbf{0.681667}                          \\
SmallKitchenAppliances         & 0.418667          & 0.658667         & \textbf{0.661333}     & 0.621333      & 0.560000                              \\
SonyAIBORobotSurface           & 0.811980          & \textbf{0.893511}& 0.835275     & \textbf{0.893511}      & 0.860233                          \\
SonyAIBORobotSurfaceII         & 0.793284          & 0.772298         & 0.766002     & 0.811123      & \textbf{0.830010}                           \\
Strawberry                     & 0.668842          & 0.649266         & 0.616639     & \textbf{0.843393}      & 0.786297                          \\
SwedishLeaf                    & 0.702400          & 0.723200         & 0.681600     & 0.806400      & \textbf{0.836800}                            \\
Symbols                        & 0.864322          & \textbf{0.954774}& \textbf{0.954774}     & 0.857286      & 0.906533                          \\
Synthetic\_control             & 0.916667          & 0.980000         & 0.980000     & 0.950000      & \textbf{0.950000}                          \\
ToeSegmentation1               & 0.574561          & \textbf{0.671053}& 0.614035     & 0.640351      & 0.653509             \\
ToeSegmentation2               & 0.546154          & \textbf{0.853846}& 0.838462     & 0.753846      & 0.746154                          \\
Trace                          & 0.580000          & \textbf{0.970000}& \textbf{0.970000}& 0.780000  & 0.800000                          \\
Two\_Patterns                  & 0.464750          & 0.989750         & \textbf{0.975000}& 0.555750  & 0.700500                            \\
TwoLeadECG                     & 0.554873          & 0.801580         & 0.811238     & \textbf{0.956102}      & 0.955224                               \\
uWaveGestureLibrary\_X         & 0.631212          & 0.706868         & 0.676438     & 0.681184      & \textbf{0.721943}                          \\
uWaveGestureLibrary\_Y         & 0.548297          & 0.564768         & 0.525405     & 0.611669      & \textbf{0.617253}                          \\
uWaveGestureLibrary\_Z         & 0.537409          & 0.604132         & 0.592406     & 0.642099      & \textbf{0.646287}                          \\
UWaveGestureLibraryAll         & 0.849525          & 0.833613         & 0.831937     & \textbf{0.920715}      & 0.911502                          \\
wafer                          & 0.654445          & 0.649416         & 0.511032     & \textbf{0.988968}      & 0.982803                          \\
Wine                           & 0.555556          & 0.574074         & 0.518519     & 0.574074      & \textbf{0.592593}                          \\
WordsSynonyms                  & 0.271160          & 0.412226         & 0.344828     & 0.474922      & \textbf{0.501567}                          \\
Worms                          & 0.215470          & 0.408840         & \textbf{0.414365}     & 0.259669      & 0.342541                        \\
WormsTwoClass                  & 0.541436          & 0.651934         & 0.591160     & 0.618785      & \textbf{0.618785}                          \\
yoga                           & 0.497000          & 0.574000         & 0.557000     & 0.631667      & \textbf{0.696667}                          \\ \hline\hline
\caption{Quantitative results of nearest centriod classification on UCR archive~\cite{chen2015ucr}.}
\label{tab:ucr_result}
\end{longtable}

\end{document}